\documentclass{article}

    \usepackage[nonatbib,preprint]{neurips_2021}

\usepackage[utf8]{inputenc} % allow utf-8 input
\usepackage[T1]{fontenc}    % use 8-bit T1 fonts
\usepackage{hyperref}       % hyperlinks
\usepackage{url}            % simple URL typesetting
\usepackage{booktabs}       % professional-quality tables
\usepackage{amsfonts}       % blackboard math symbols
\usepackage{nicefrac}       % compact symbols for 1/2, etc.
\usepackage{microtype}      % microtypography
\usepackage{times}
\usepackage{epsfig}
\usepackage{graphicx}
\usepackage{amsmath}
\usepackage{amssymb}
\usepackage{sidecap}    
\usepackage{multirow}
\usepackage{subcaption}

\usepackage{pifont}
\usepackage{overpic}
\newcommand{\cmark}{\ding{51}}%
\newcommand{\xmark}{\ding{55}}%

\newcommand{\highlight}[1]{\textcolor{black}{\textbf{#1}}}

\usepackage{float} 
\usepackage[dvipsnames]{xcolor}
\definecolor{mygreen}{RGB}{0,100,0}
\definecolor{myblue}{RGB}{0,0,240}
\definecolor{myred}{RGB}{200,0,0}

\newcommand{\myPara}[1]{\noindent\textbf{#1:}}

\newcommand{\nameofmethod}{LV-ViT}

\newcommand{\cmarkgreen}{\textcolor{mygreen}{\ding{51}}}%

\def \pzo {\phantom{0}} 
\def \dzo {\phantom{00}}

\def \OURS {LV-ViT}

\title{All Tokens Matter: Token Labeling for Training Better Vision Transformers}

\author{Zihang Jiang$^{1}$\thanks{Work done as an intern at ByteDance AI Lab.} \And Qibin Hou$^1$ \And Li Yuan$^1$ \And Daquan Zhou$^1$ \And Yujun Shi$^1$  \AND Xiaojie Jin$^2$ \And Anran Wang$^2$ \And Jiashi Feng$^1$ \AND
$^1$National University of Singapore  \And  $^2$ByteDance \And
{\tt\small \{jzh0103,andrewhoux,ylustcnus,zhoudaquan21,shiyujun1016\}@gmail.com} \\
{\tt\small xjjin0731@gmail.com, anran.wang@bytedance.com, elefjia@nus.edu.sg}
}

\begin{document}

\maketitle

\begin{abstract}
In this paper, we present token labeling\textemdash a new training objective 
for training high-performance vision transformers (ViTs).
Different from the standard training objective of ViTs that computes the classification loss on an additional trainable class token,  
our proposed one takes advantage of all the image patch tokens to compute the training loss in a dense manner.
Specifically, token labeling reformulates the image classification problem into multiple token-level recognition problems and 
assigns each patch token with an individual location-specific supervision  generated by a machine annotator.
Experiments show that token labeling can clearly and consistently improve the performance of various ViT models across a wide spectrum.
For a vision transformer with 26M learnable parameters serving as an example, with token labeling, the model can achieve 84.4\% Top-1 accuracy on ImageNet.
The result can be further increased to 86.4\% by slightly scaling the model size up to 150M, delivering the minimal-sized model among previous models (250M+) reaching 86\%.
We also show that token labeling can clearly improve the generalization capability of the pretrained models  on downstream tasks with dense prediction, such as semantic segmentation.
Our code and all the training details are publicly available at \url{https://github.com/zihangJiang/TokenLabeling}.

\end{abstract}

\section{Introduction} \label{sec:introduction}

Transformers \cite{vaswani2017attention} have achieved great performance for almost all the natural language processing (NLP) tasks over the past years \cite{brown2020language,devlin2018bert,liu2019roberta}.
Motivated by such success, recently, many researchers attempt to build transformer models for vision tasks, and their encouraging results have shown the great potential of transformer based models for image classification \cite{chen2021crossvit,dosovitskiy2020image,liu2021swin,touvron2020training,wang2021pyramid,yuan2021tokens}, 
especially the strong benefits of the self-attention mechanism in building long-range dependencies between pairs of input tokens.

\definecolor{mygray}{gray}{0.35}

\begin{figure}[h]
    \centering
    \tiny
    \begin{overpic}[width=\linewidth]{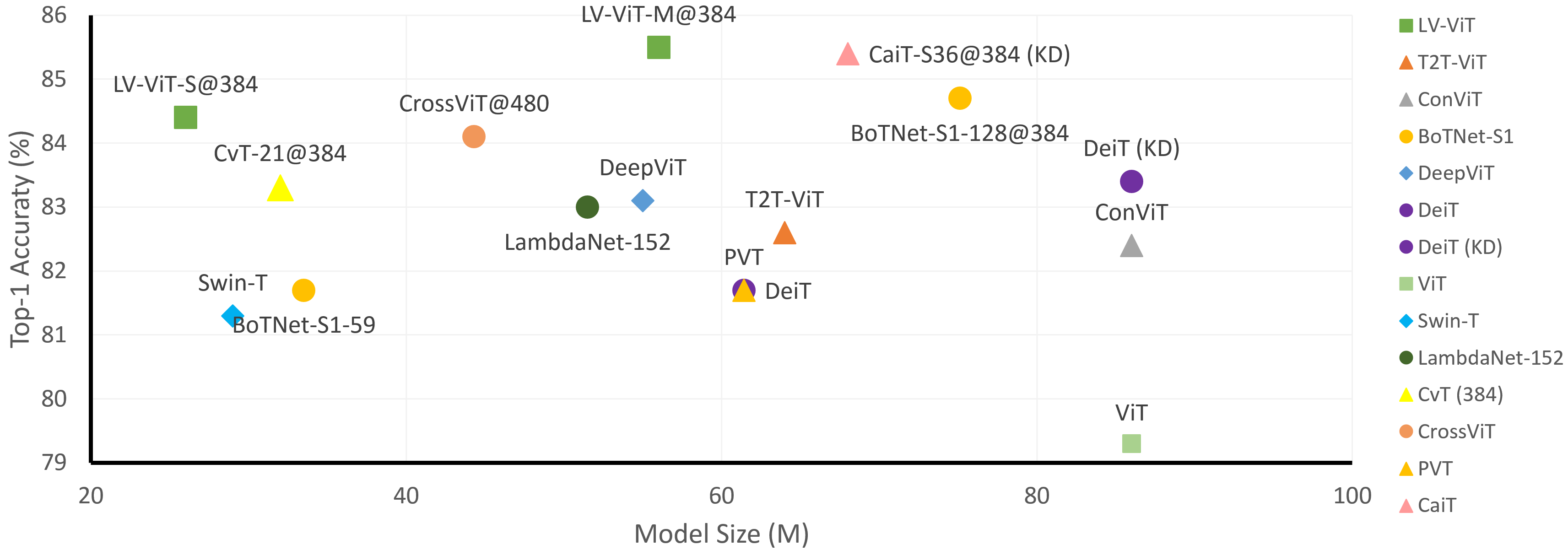}
    \end{overpic}
    \caption{Comparison between the proposed \nameofmethod{} and other recent works based on vision transformers, including T2T-ViT~\cite{yuan2021tokens}, ConViT~\cite{d2021convit}, BoTNet~\cite{srinivas2021bottleneck}, DeepViT~\cite{zhou2021deepvit}, DeiT~\cite{touvron2020training}, ViT~\cite{dosovitskiy2020image}, Swin Transformer~\cite{liu2021swin}, LambdaNet~\cite{bello2021lambdanetworks}, CvT~\cite{wu2021cvt}, CrossViT~\cite{chen2021crossvit}, PVT~\cite{wang2021pyramid}, CaiT~\cite{touvron2021going}.
    Note that we only show models whose model sizes are under 100M. As can be seen, our \nameofmethod{}
    achieves the best results using the least amount of learnable parameters. The default test resolution
    is $224\times224$ unless specified after @.}
    \label{fig:res_fig}
\end{figure}

Despite the importance of gathering long-range dependencies, recent work on local data augmentation~\cite{zhong2020random} has demonstrated that well modeling and leveraging local information for image classification would avoid biasing the model towards skewed and non-generalizable patterns and substantially improve the model performance.
However, recent vision transformers normally utilize class tokens that aggregate global information to predict the output class while neglecting the role of other patch tokens that encode rich information on their respective local image patches.

In this paper, we present a new training objective for vision transformers, termed \emph{token labeling}, that takes advantage of both the patch tokens and the class tokens.
Our method takes a $K$-dimensional score map generated by a machine annotator as supervision to supervise all the tokens in a dense manner,
where $K$ is the number of categories for the target dataset.
In this way, each patch token is explicitly associated with an individual location-specific supervision indicating the existence of the target objects inside the corresponding image patch,
so as to improve the object grounding and recognition capabilities of vision transformers with negligible computation overhead.
To the best of our knowledge, this is the first work demonstrating that dense supervision is beneficial to vision transformers in image classification.

According to our experiments, utilizing the proposed token labeling objective can clearly boost the performance of vision transformers.
As shown in Figure~\ref{fig:res_fig}, our model, named \nameofmethod{}, with 56M parameters, yields
85.4\% top-1 accuracy on ImageNet \cite{deng2009imagenet}, behaving better than
all the other transformer-based models having no more than 100M parameters.
When the model size is scaled up to 150M, the result can be further improved to 86.4\%.
In addition, we have empirically found that the pretrained models with token labeling are also beneficial to downstream tasks with dense prediction, 
such as semantic segmentation.

\section{Related Work}

Transformers~\cite{vaswani2017attention} refer to the models that entirely rely on the self-attention mechanism to build global dependencies, which are originally designed for natural language processing tasks.
Due to their strong capability of capturing spatial information, transformers have also been successfully applied to a variety of vision problems, including low-level vision tasks like image enhancement~\cite{chen2020pre,yang2020learning}, as well as more challenging tasks
such as image classification~\cite{chen2020generative, dosovitskiy2020image}, object 
detection~\cite{carion2020end,dai2020up,zheng2020end,zhu2020deformable},
segmentation~\cite{chen2020pre,sun2020rethinking,wang2020end} and image generation~\cite{parmar2018image}. 
Some works also extend transformers for video and 3D point cloud processing~\cite{zeng2020learning,zhao2020point,zhou2018end}.

\begin{figure*}[t]
    \centering
    \small
    \includegraphics[width=0.9\linewidth]{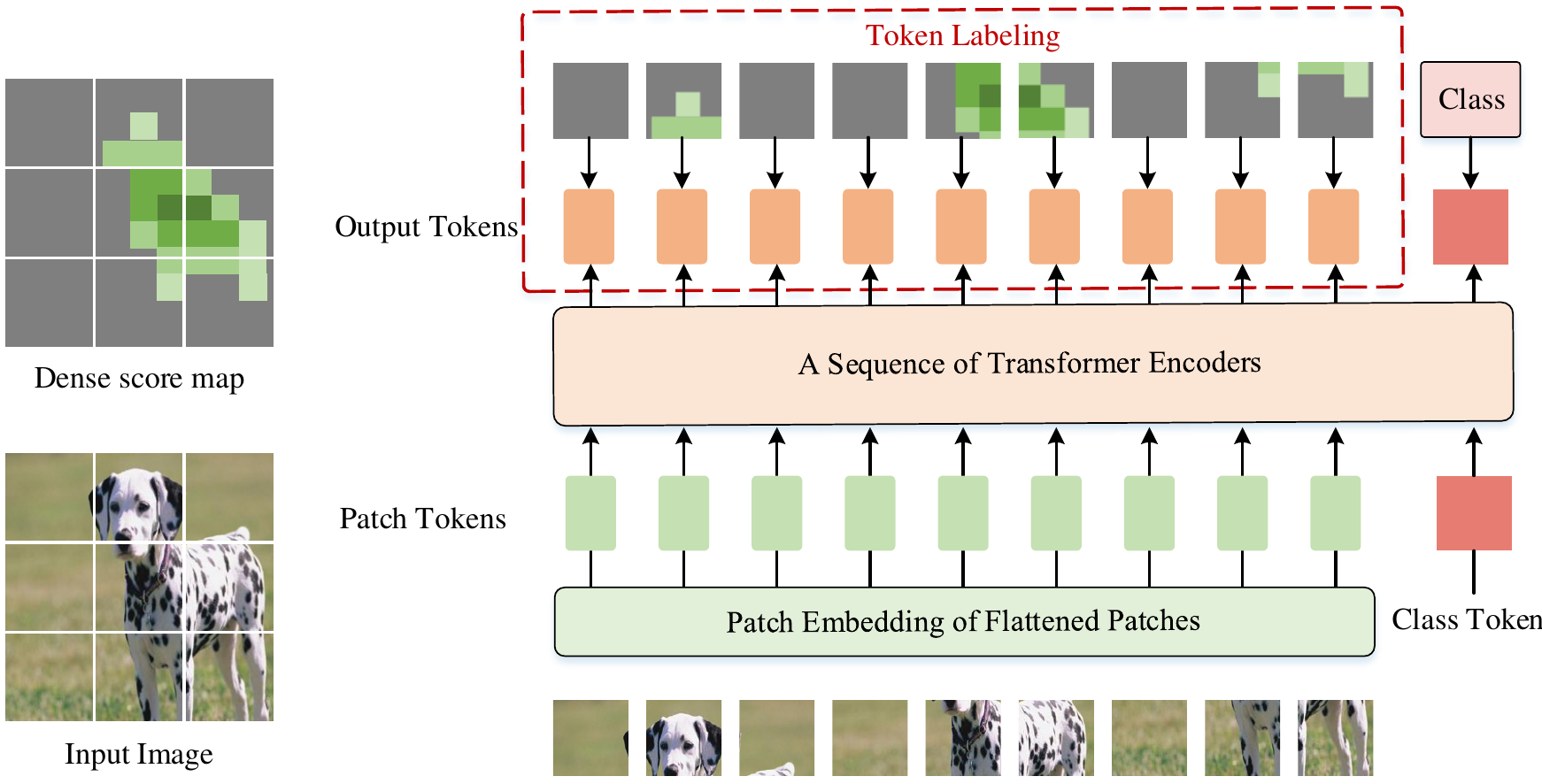}
    \caption{Pipeline of training vision transformers with token labeling. Other than utilizing the class token (pink rectangle), we also take advantage of all the output patch tokens (orange rounded rectangle) by assigning each patch token an individual location-specific prediction generated by a machine annotator \cite{brock2021high} as supervision (see the part in the red dash rectangle).
    Our proposed token labeling method can be treated as an auxiliary objective to provide each patch token the local details that aid vision transformers to more accurately locate and recognize the target objects. Note that the traditional vision transformer training does not include the red dash rectangle part.}
    \label{fig:arch}
\end{figure*}

Vision Transformer (ViT) is one of the earlier attempts that achieved state-of-the-art performance on ImageNet classification, using pure transformers as basic building blocks.
However, ViTs need pretraining on very large datasets, such as ImageNet-22k and JFT-300M,
and huge computation resources to achieve comparable performance to ResNet~\cite{he2016deep} with a similar model size trained on ImageNet.
Later, DeiT~\cite{touvron2020training} manages to tackle the data-inefficiency problem by simply adjusting the network architecture and adding an additional token along with the class token for Knowledge Distillation~\cite{hinton2015distilling, yuan2020revisiting} to improve model performance.

Some recent works~\cite{chen2021crossvit,han2021transformer,wu2021cvt,yuan2021tokens} also attempt to introduce the local dependency into vision transformers by modifying the patch embedding block or the transformer block or both, leading to significant performance gains.
Moreover, there are also some works~\cite{heo2021rethinking,liu2021swin,wang2021pyramid} adopting a pyramid structure to reduce the overall computation while maintaining the model's ability to capture low-level features.

Unlike most aforementioned works that design new transformer blocks or transformer architectures, we attempt to improve vision transformers by studying the role of patch tokens that embed rich local information inside image patches.
We show that by slightly tuning the structure of vision transformers and employing the proposed token labeling objective,
we can achieve strong baselines for transformer models at different model size levels.

\section{Token Labeling Method}

In this section, we first briefly review the structure of the vision transformer \cite{dosovitskiy2020image}
and then describe the proposed training objective---\emph{token labeling}.

\subsection{Revisiting Vision Transformer}

A typical vision transformer \cite{dosovitskiy2020image} first decomposes a fixed-size input image into a sequence of small patches.
Each small patch is mapped to a feature vector, or called a token, by projection with a linear layer.
Then, all the tokens combined with an additional learnable class token for classification score prediction are sent into a stack of transformer blocks for feature encoding.

In loss computing, the class token from the output tokens of the last transformer block is usually selected and sent into a linear layer for the classification score prediction.
Mathematically, given an image $I$, denote the output of the last transformer block as $[X^{cls},X^1,...,X^N]$,
where $N$ is the total number of patch tokens, and $X^{cls}$ and $X^1,...,X^N$ correspond to the class token
and the patch tokens, respectively.
The classification loss for image $I$ can be written as
\begin{equation} \label{eqn:cross_entropy}
    L_{cls} = H(X^{cls},y^{cls}),
\end{equation}
where $H(\cdot, \cdot)$ is the softmax cross-entropy loss and $y^{cls}$ is the class label.

\subsection{Token Labeling} \label{sec:token_labeling}

The above classification problem only adopts an image-level label as supervision whereas it neglects the rich information embedded in each image patch.
In this subsection, we present a new training objective---\emph{token labeling}---that takes advantage of the complementary information between the patch tokens and the class tokens.

\myPara{Token Labeling} Different from the classification loss as formulated in Eqn.~(\ref{eqn:cross_entropy})
that measures the distance between the single class token (representing the whole input image) and the corresponding image-level label, 
token labeling emphasizes the importance of all output tokens and advocates that each output token should be associated with an individual location-specific label.
Therefore, in our method, the ground truth for an input image involves not only a single $K$-dimensional vector
$y^{cls}$ but also a $K \times N$ matrix or called a $K$-dimensional score map as represented by $[y^1,...,y^N]$,
where $N$ is the number of the output patch tokens.

Specifically, we leverage a dense score map for each training image and 
use the cross-entropy loss between each output patch token and 
the corresponding aligned label in the dense score map as an auxiliary loss
at the training phase.
Figure~\ref{fig:arch} provides an intuitive interpretation.
Given the output patch tokens $X^1,...,X^N$ and the corresponding labels $[y^1,...,y^N]$,
the token labeling objective can be defined as
\begin{equation}
    L_{tl}=\frac{1}{N}\sum_{i=1}^N H(X^i,y^i).
\end{equation}
Recall that $H$ is the cross-entropy loss.
Therefore, the total loss function can be written as
\begin{align}
    L_{total} &= H(X^{cls},y^{cls}) + \beta \cdot L_{tl}, \\
              &= H(X^{cls},y^{cls}) + \beta \cdot \frac{1}{N}\sum_{i=1}^N H(X^i,y^i),
\end{align}
where $\beta$ is a hyper-parameter to balance the two terms.
In our experiment, we empirically set it to 0.5.

\myPara{Advantages} Our token labeling offers the following advantages. 
\emph{First of all}, unlike knowledge distillation methods that require a teacher model to generate supervision labels online, 
token labeling is a cheap operation. 
The dense score map can be generated by a pretrained model in advance
(e.g., EfficientNet \cite{tan2019efficientnet} or NFNet \cite{brock2021high}).
During training, we only need to crop the score map and perform interpolation to make it aligned with the cropped image in the spatial coordinate.
Thus, the additional computations are negligible.
\emph{Second}, rather than utilizing a single label vector as supervision as done in most classification models 
and the ReLabel strategy \cite{yun2021relabel},
we also harness score maps to supervise the models in a dense manner and thereby
the label for each patch token provides location-specific information, 
which can aid the training models to easily discover the target objects and improve the recognition accuracy.
\emph{Last but not the least}, as dense supervision is adopted in training, 
we found that the pretrained models with token labeling benefit downstream tasks with dense prediction, like semantic segmentation.

\begin{SCfigure}
    % \centering
    % \small
    \caption{Comparison between CutMix~\cite{yun2019cutmix} (\textbf{Left}) and our proposed MixToken (\textbf{Right}).
    CutMix is operated on the input images. This results in patches containing mixed regions from the two images
    (see the patches enclosed by red bounding boxes).
    Differently, MixToken targets at mixing tokens after patch embedding. This enables each token
    after patch embedding to have clean content as shown in the right part of this figure. The detailed advantage
    of MixToken can be found in Sec.~\ref{sec:ablation}.}
    \includegraphics[width=0.45\textwidth]{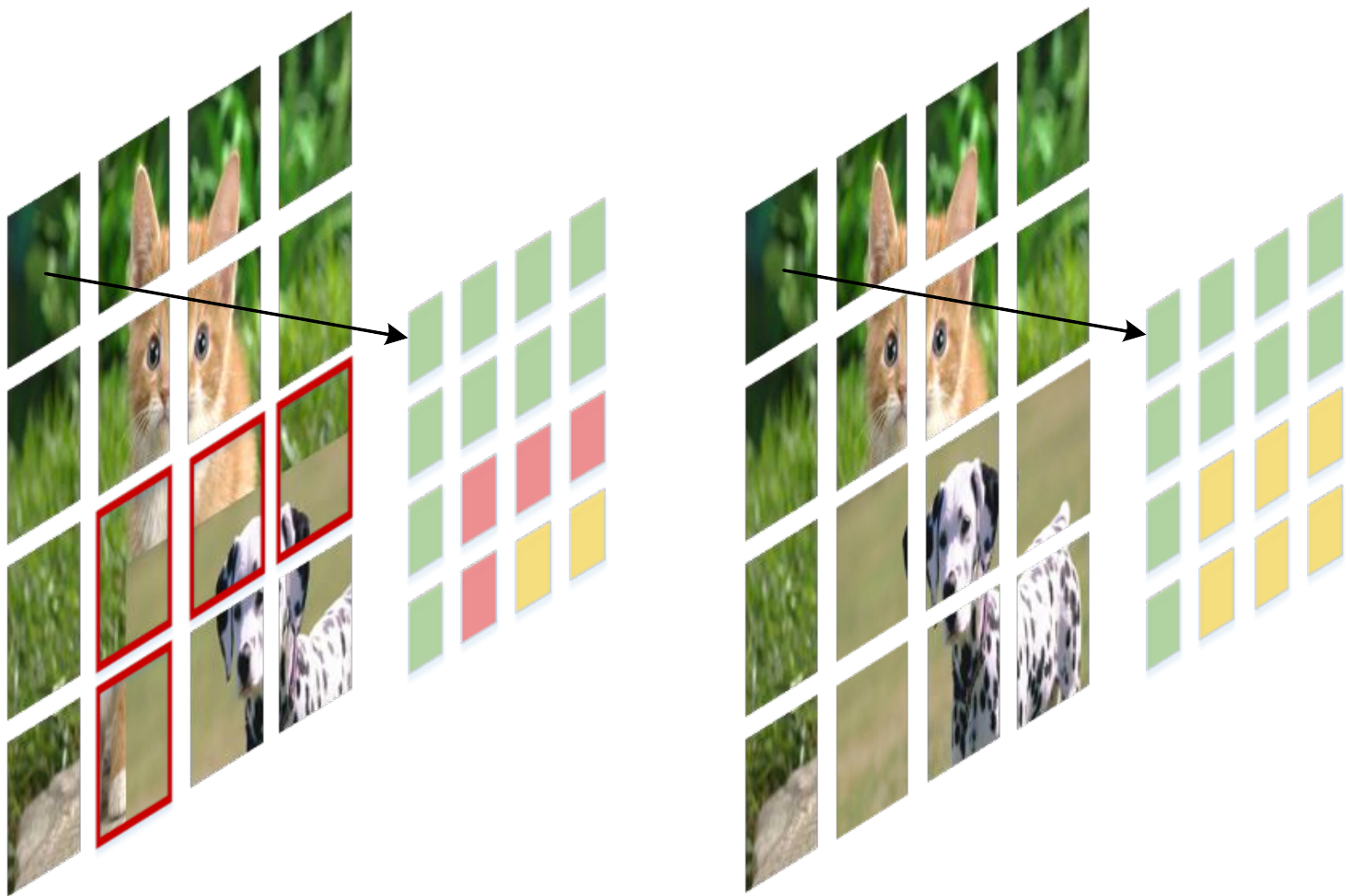}
    \label{fig:mix_token}
\end{SCfigure}

\subsection{Token Labeling with MixToken}

While training vision transformer, previous studies~\cite{touvron2020training, yuan2021tokens} 
have shown that augmentation methods, like MixUp~\cite{zhang2017mixup} and CutMix~\cite{yun2019cutmix},
can effectively boost the performance and robustness of the models.
However, vision transformers rely on patch-based tokenization to map each input image to a sequence of tokens
and our token labeling strategy also operates on patch-based token labels.
If we apply CutMix directly on the raw image, some of the resulting patches may contain 
content from two images, leading to mixed regions within a small patch as shown in Figure~\ref{fig:mix_token}.
When performing token labeling, it is difficult to assign each output token a clean and correct label.
Taking this situation into account, we rethink the CutMix augmentation method and present MixToken, 
which can be viewed as a modified version of CutMix operating on the tokens after patch embedding
as illustrated in the right part of Figure~\ref{fig:mix_token}. 

To be specific, for two images denoted as $I_1,I_2$ and their corresponding token labels $Y_1=[y_1^1,...,y^N_1]$
as well as $Y_2=[y^1_2,...,y^N_2]$, we first feed the two images into the patch embedding module to 
tokenize each as a sequence of tokens, resulting in $T_1=[t^1_1,...,t^N_1]$ and $T_2=[t^1_2,...,t^N_2]$. 
Then, we produce a new sequence of tokens by applying MixToken using a binary mask $M$ as follows:
\begin{equation}
    \hat{T}=T_1\odot M + T_2 \odot (1-M),
\end{equation}
where $\odot$ is element-wise multiplication. We use the same way to generate the mask $M$
as in~\cite{yun2019cutmix}.
For the corresponding token labels, we also mix them using the same mask $M$: % and corresponding ratio $\lambda$
\begin{equation}
    \hat{Y}=Y_1\odot M + Y_2 \odot (1-M). 
\end{equation}
The label for the class token can be written as
\begin{equation}
    \hat{y^{cls}}= \Bar{M} y_1^{cls} + (1 - \bar{M}) y_2^{cls},
\end{equation}
where $\Bar{M}$ is the average of all element values of $M$.

\section{Experiments}

\subsection{Experiment Setup}

We evaluate our method on the ImageNet \cite{deng2009imagenet} dataset.
All experiments are built and conducted upon \texttt{PyTorch}~\cite{paszke2019pytorch} and the \texttt{timm}~\cite{rw2019timm} library.
We follow the standard training schedule and train our models on the ImageNet dataset for 300 epochs.
Besides normal augmentations like CutOut~\cite{zhong2020random} and RandAug~\cite{cubuk2020randaugment}, 
we also explore the effect of applying MixUp~\cite{zhang2017mixup} and CutMix~\cite{yun2019cutmix} together
with our proposed token labeling.
Empirically, we have found that using MixUp together with token labeling brings no benefit to the performance, and thus we do not apply it in our experiments.

For optimization, by default, we use the AdamW optimizer~\cite{loshchilov2017decoupled} 
with a linear learning rate scaling strategy $lr = 10^{-3} \times \frac{batch\_size}{640}$ and $5\times 10^{-2}$
weight decay rate. 
For Dropout regularization, we observe that for small models, using Dropout hurts the performance.
This has also been observed in a few other works related to training vision transformers~\cite{touvron2020training, touvron2021going, yuan2021tokens}.
As a result, we do not apply Dropout~\cite{srivastava2014dropout} and use 
Stochastic Depth~\cite{huang2016deep} instead.
More details on hyper-parameters and finetuning can  be found in our supplementary materials.

We use the NFNet-F6~\cite{brock2021high} trained on ImageNet with an $86.3\%$ Top-1 accuracy 
as the machine annotator to generate dense score maps for the ImageNet dataset,
yielding a 1000-dimensional score map for each image for training.
The score map generation procedure is similar to~\cite{yun2021relabel}, but 
we limit our experiment setting by training all models from scratch on ImageNet without extra data support, such as JFT-300M and ImageNet-22K.
This is different from the original ReLabel paper \cite{yun2021relabel}, 
in which the EfficientNet-L2 model pretrained on JFT-300M is used.
%
% Thus we choose NFNet-F6~\cite{brock2021high} trained on ImageNet as our machine annotator to provide dense score map. 
%
The input resolution for NFNet-F6 is $576 \times 576$, and the dimension of 
the corresponding output score map for each image 
is $L\in \mathbb{R}^{18\times 18\times 1000}$. 
During training, the target labels for the tokens are generated by applying RoIAlign~\cite{he2017mask} on the corresponding score map.
In practice, we only store the top-5 score maps for each position in half-precision to
save space as storing the entire score maps for all the images results in 2TB storage.
In our experiment, we only need 10GB of storage to store all the score maps.

\subsection{Ablation Analysis} \label{sec:ablation}

\myPara{Model Settings}
The default settings of the proposed LV-ViT are given in Table~\ref{tab:model_size}, where both token labeling and MixToken are used. 
A slight architecture modification to ViT \cite{dosovitskiy2020image} is that we replace the patch embedding module with a 4-layer convolution to better 
tokenize the input image and integrate local information.
Detailed ablation about patch embedding can be found in our supplementary materials.
As can be seen, our LV-ViT-T with only 8.5M parameters can already achieve
a top-1 accuracy of 79.1\% on ImageNet.
Increasing the embedding dimension and network depth can further boost the performance.
More experiments compared to other methods can be found in Sec.~\ref{sec:comp_others}.
In the following ablation experiments, we will set our LV-ViT-S as baseline and show
the advantages of the proposed token labeling and MixToken methods.

\begin{table}[t]
  \centering
  \small
  \setlength\tabcolsep{1.9mm}
  \renewcommand\arraystretch{1.0}
  \caption{Performance of the proposed \nameofmethod{} with different model sizes. Here, `depth' denotes
  the number of transformer blocks used in different models. By default, the test resolution is set
  to $224 \times 224$ except the last one which is $288 \times 288$.}
  \label{tab:model_size}
  \begin{tabular}{ccccccccc} \toprule[0.5pt]
    Name & Depth & Embed dim. & MLP Ratio & \#Heads & \#Parameters & Resolution & Top-1 Acc. (\%) \\ \midrule[0.5pt] \midrule[0.5pt]
    \nameofmethod{}-T & 12 & 240 & 3.0 & 4  & 8.5M & 224x224 & 79.1\\
    \nameofmethod{}-S & 16 & 384 & 3.0 & 6  & 26M  & 224x224 & 83.3\\ 
    \nameofmethod{}-M & 20 & 512 & 3.0 & 8  & 56M  & 224x224 & 84.1\\
    \nameofmethod{}-L & 24 & 768 & 3.0 & 12 & 150M & 288x288 & \highlight{85.3}\\
    \bottomrule[0.5pt]
  \end{tabular}
\end{table}

\myPara{MixToken}
We use MixToken as a substitution for CutMix while applying token labeling.
Our experiments show that MixToken performs better than CutMix for token-based transformer models.
As shown in Table~\ref{tab:abl_mixtoken}, when training with the original ImageNet labels, using MixToken is $0.1\%$ higher than using CutMix. 
When using the ReLabel supervision, we can also see an advantage of $0.2\%$ over the CutMix baseline.
Combining with our token labeling, the performance can be further raised to $83.3\%$.

\begin{table}[h]
  \begin{minipage}[t]{0.45\textwidth}
  \centering
  \small
  \setlength\tabcolsep{2mm}
  \renewcommand\arraystretch{1}
  \caption{Ablation on the proposed MixToken and token labeling augmentations. We also show results with either the ImageNet hard label and the ReLabel \cite{yun2021relabel} as supervision.}
  
  \label{tab:abl_mixtoken}
  \begin{tabular}{cccccc} \toprule[0.5pt]
    Aug. Method & Supervision &Top-1 Acc. \\ \midrule[0.5pt] \midrule[0.5pt]
     MixToken & Token labeling & \highlight{83.3} \\
     MixToken & ReLabel & 83.0\\
     CutMix & ReLabel & 82.8\\
     Mixtoken & ImageNet Label & 82.5\\
     CutMix & ImageNet Label & 82.4\\
    \bottomrule[0.5pt]
  \end{tabular}
  \end{minipage}
  \hfill
  \begin{minipage}[t]{0.52\textwidth}
  \centering
  \small
  \setlength\tabcolsep{1.5mm}
  \renewcommand\arraystretch{1}
  \caption{Ablation on different widely-used data augmentations. We have empirically found our proposed
  MixToken performs even better than the combination of MixUp and CutMix in vision transformers.}
  \label{tab:abl_aug}
  \begin{tabular}{cccccc} \toprule[0.5pt]
    MixToken & MixUp & CutOut & RandAug & Top-1 Acc. \\ \midrule[0.5pt] \midrule[0.5pt]
     \checkmark & \xmark  &\checkmark &\checkmark  &  \highlight{83.3} \\
     \xmark & \xmark  &\checkmark &\checkmark  &   81.3\\
     \checkmark & \checkmark  &\checkmark &\checkmark  &  83.1 \\
     \checkmark & \xmark  &\xmark &\checkmark  &  83.0\\ 
     \checkmark & \xmark  &\checkmark &\xmark  & 82.8\\
    \bottomrule[0.5pt]
  \end{tabular}
  \end{minipage}
\end{table}

\myPara{Data Augmentation}
Here, we study the compatibility of MixToken with other augmentation techniques,
such as MixUp~\cite{zhang2017mixup}, CutOut~\cite{zhong2020random} and RandAug~\cite{cubuk2020randaugment}.
The ablation results are shown in Table~\ref{tab:abl_aug}.
We can see when all the four augmentation methods are used, a top-1 accuracy of 
83.1\% is achieved.
Interestingly, when the MixUp augmentation is removed, the performance can be improved to 83.3\%.
This may be explained as, using MixToken and MixUp at the same time would bring 
too much noise in the label, and consequently cause confusion of the model.
Moreover, the CutOut augmentation, which randomly erases some parts of the image, 
is also effective and removing it brings a performance drop of $0.3\%$.
Similarly, the RandAug augmentation also contributes to the performance and using it brings an improvement of $0.5\%$.

\begin{figure*}[h]
    \centering
    \scriptsize
    \includegraphics[width=0.49\linewidth]{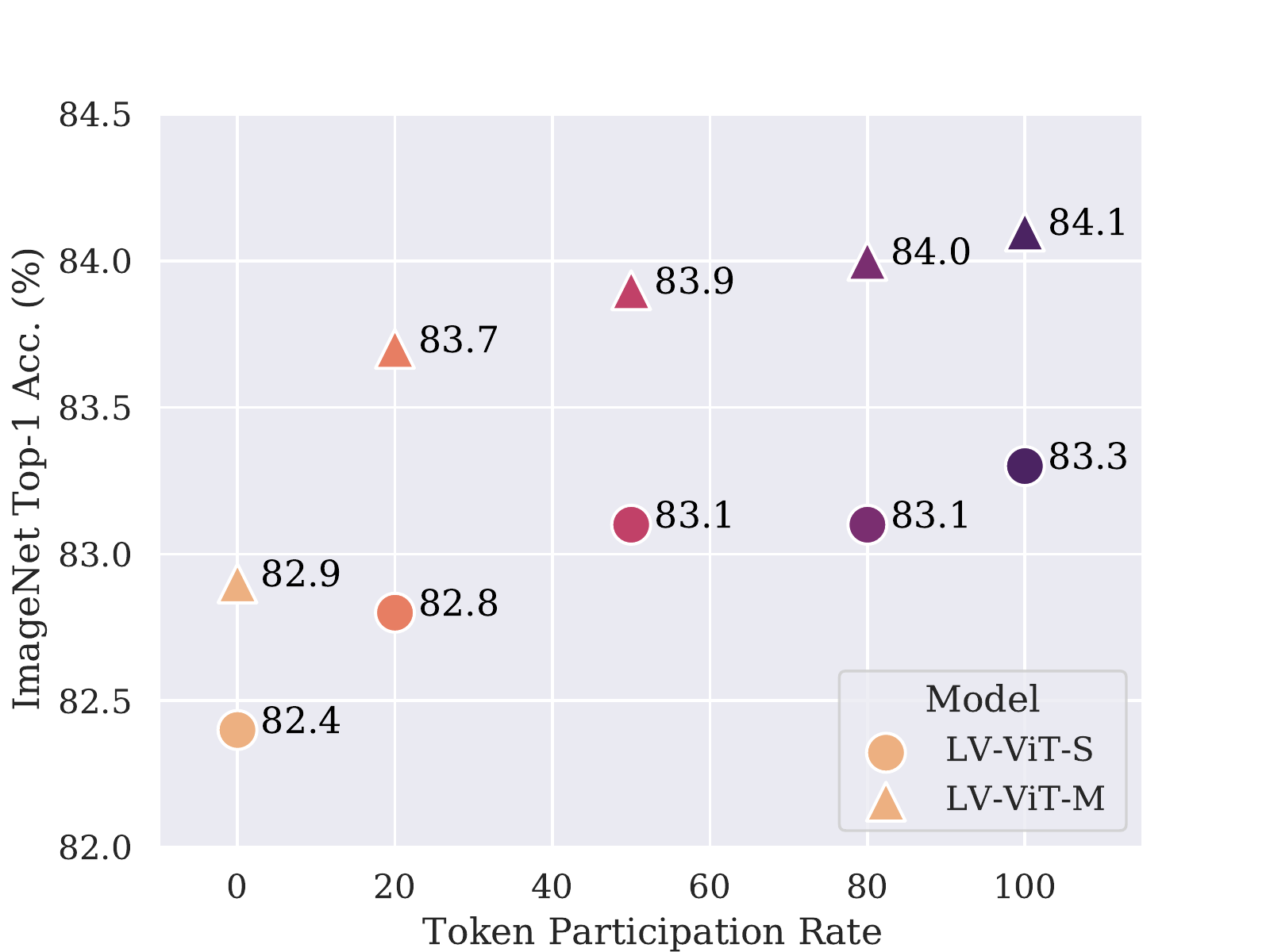}
    \includegraphics[width=0.49\linewidth]{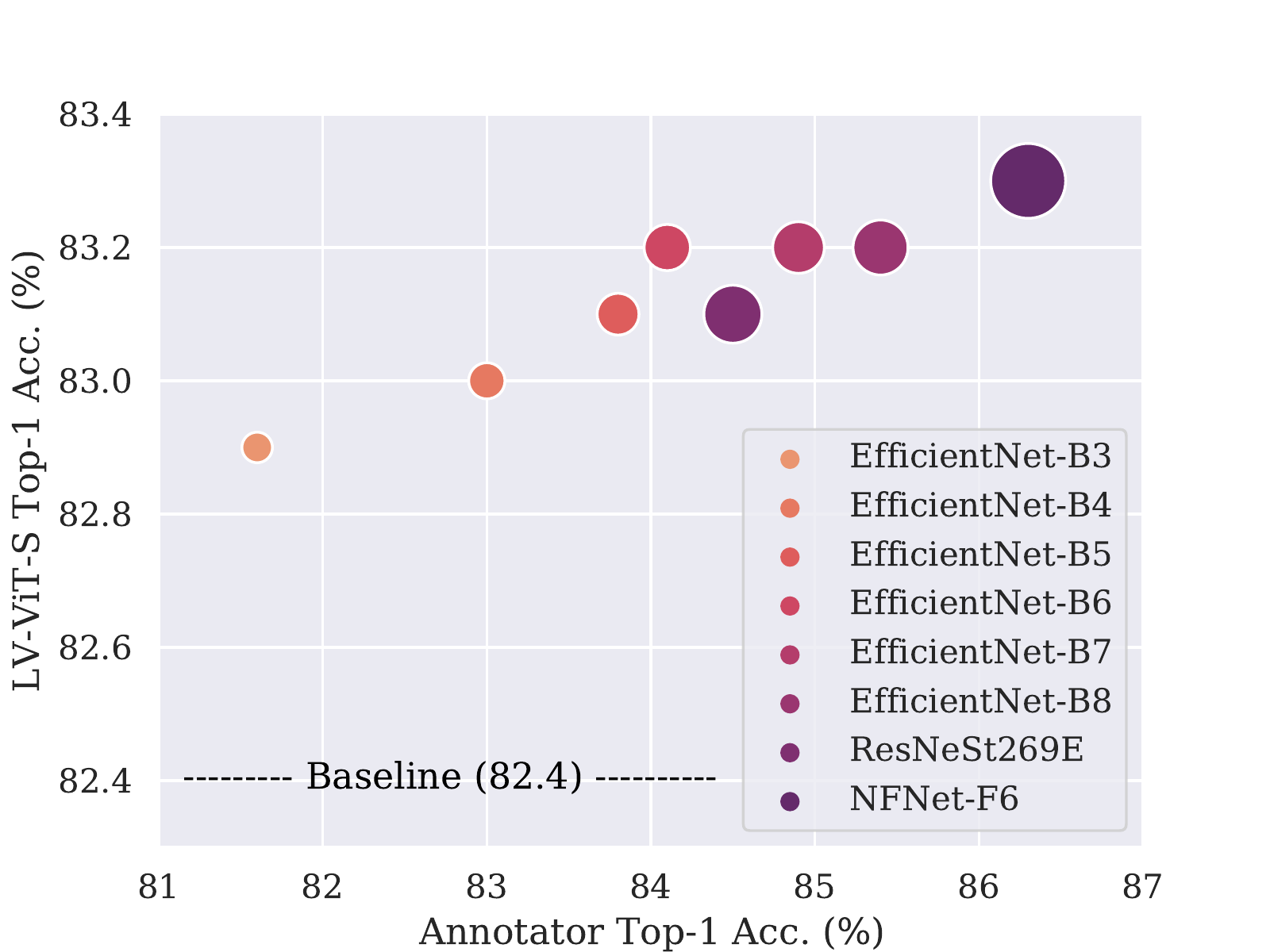}
    % \begin{overpic}[width=0.49\linewidth]{figures/annotator.png}
    %     \put(14.5, 12.6){- - - - - baseline (82.4) - - - - -}
    % \end{overpic}
    % \includegraphics[width=0.49\linewidth]{figures/annotator.png}
    \caption{\textbf{Left}: LV-ViT ImageNet Top-1 Accuracy w.r.t. the token participation rate while applying token labeling. Token participation rate indicates the percentage of patch tokens involved in computing the token labeling loss.
    This experiment reflects that all tokens matter for vision transformers. 
    \textbf{Right}: LV-ViT-S ImageNet Top-1 Accuracy w.r.t. different annotator models. The point size indicates the parameter number of the annotator model. Clearly, our token labeling objective is robust to different annotator models.}
    \label{fig:part_anno}
\end{figure*}

\myPara{All Tokens Matter}
To show the importance of involving all tokens in our token labeling method, 
we attempt to randomly drop some tokens and use the remaining ones for computing the token labeling loss.
The percentage of the remaining tokens is denoted as Token Participation Rate. 
As shown in Figure~\ref{fig:part_anno} (Left), we conduct experiments on two models: LV-ViT-S and LV-ViT-M.
As can be seen, using only $20\%$ of the tokens to compute the token labeling loss decreases the performance ($-0.5\%$ for LV-ViT-S and $-0.4\%$ for LV-ViT-M).
Involving more tokens for loss computation consistently leads to better performance.
Since involving all tokens brings negligible computation cost and gives the best performance, 
we always set the token participation rate as $100\%$ in the following experiments.

\myPara{Robustness to Different Annotators} 
To evaluate the robustness of our token labeling method, 
we use different pretrained CNNs, including EfficientNet-{B3,B4,B5,B6,B7,B8}~\cite{tan2019efficientnet}, NFNet-F6~\cite{brock2021high} and ResNest269E~\cite{zhang2020resnest}, as annotator models to provide dense supervision.
Results are shown in the right part of Figure~\ref{fig:part_anno}.
We can see that, even if we use an annotator with relatively lower performance, 
such as EfficientNet-B3 whose Top-1 accuracy is 81.6\%, it can still provide multi-label location-specific supervision and help improve the performance of our LV-ViT-S model.
Meanwhile, annotator models with better performance can provide more accurate supervision, bringing even better performance, as 
stronger annotator models can generate better token-level
labels.
The largest annotator NFNet-F6~\cite{brock2021high}, which has the best performance of $86.3\%$, allows us to achieve the best result for LV-ViT-S, which is $83.3\%$. 
In addition, we also attempt to use a better model, EfficientNet-L2 pretrained on JFT-300M as described in~\cite{yun2021relabel} which has $88.2\%$ Top-1 ImageNet accuracy, as our annotator.
The performance of LV-ViT-S can be further improved to $83.5\%$.
However, to fairly compare with the models without extra training data, we only report results based on dense supervision produced by NFNet-F6~\cite{brock2021high} 
that uses only ImageNet training data.

\begin{figure*}[h]
    \centering
    \small
    \begin{tabular}{lll}
    \includegraphics[trim={0.4cm 0.48cm 1.3cm 1cm}, clip,width=0.31\linewidth]{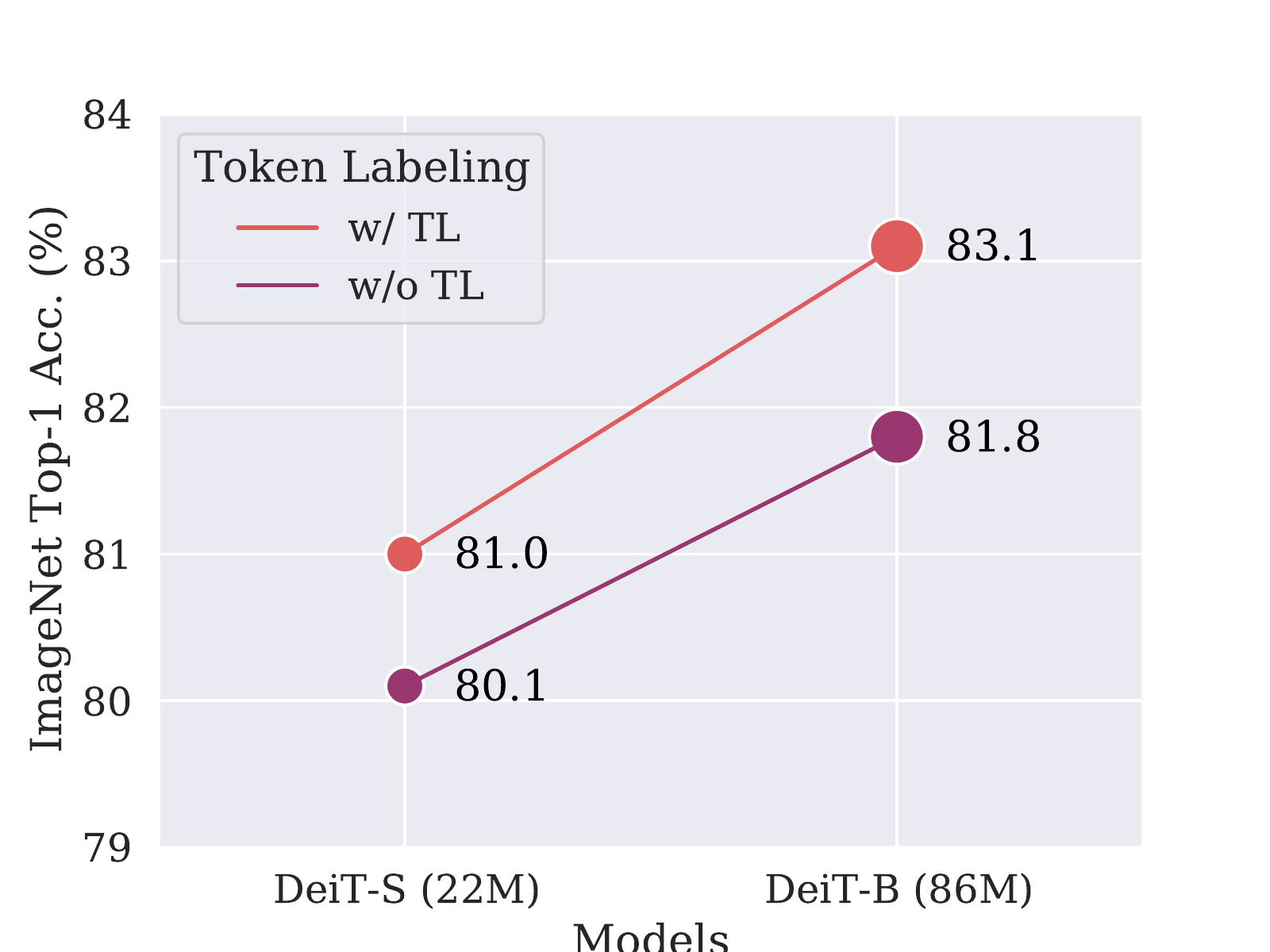} &
    \includegraphics[trim={0.4cm 0.48cm 1.3cm 1cm}, clip,width=0.31\linewidth]{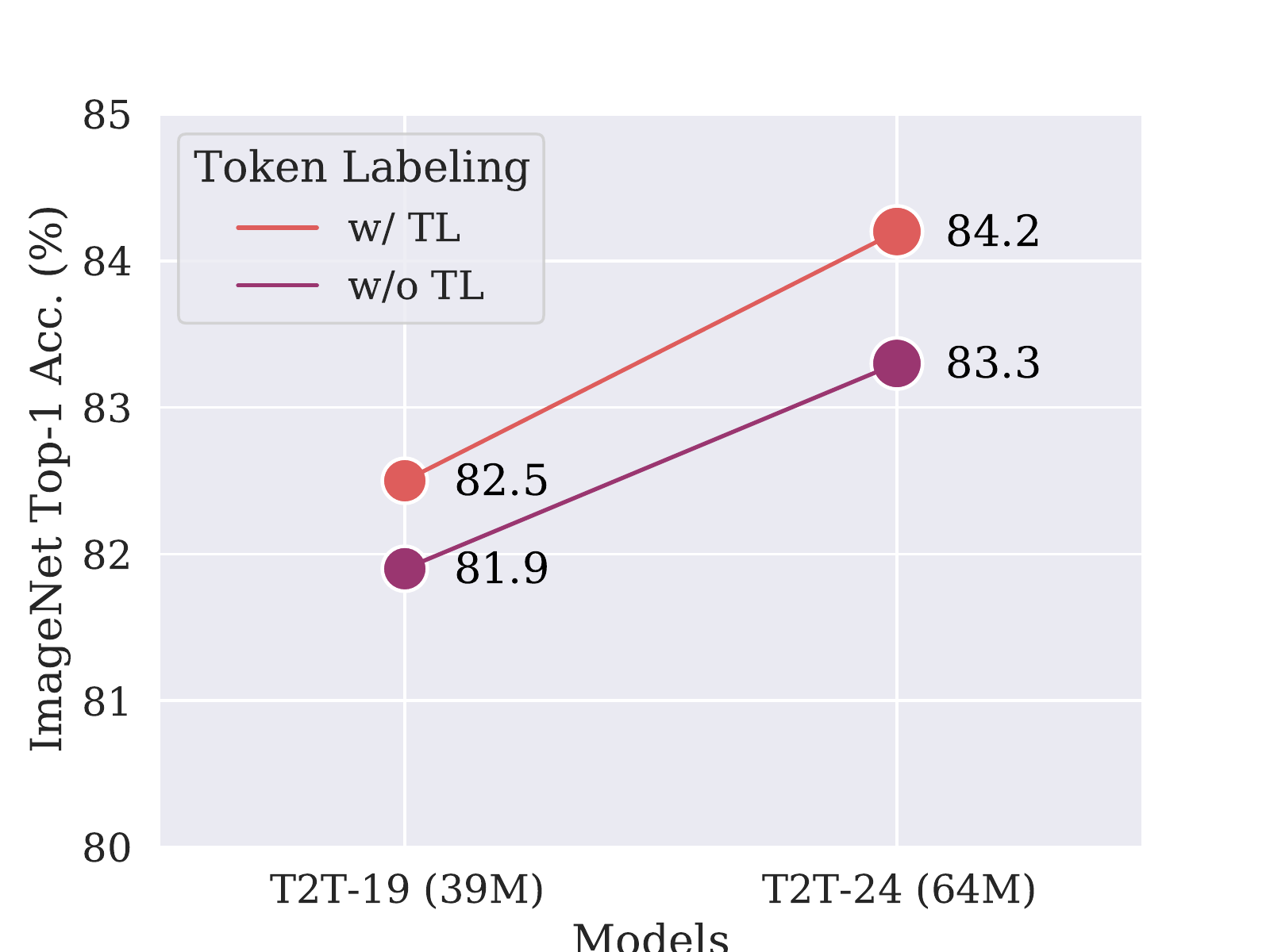} &
    \includegraphics[trim={0.4cm 0.48cm 1.3cm 1cm}, clip,width=0.31\linewidth]{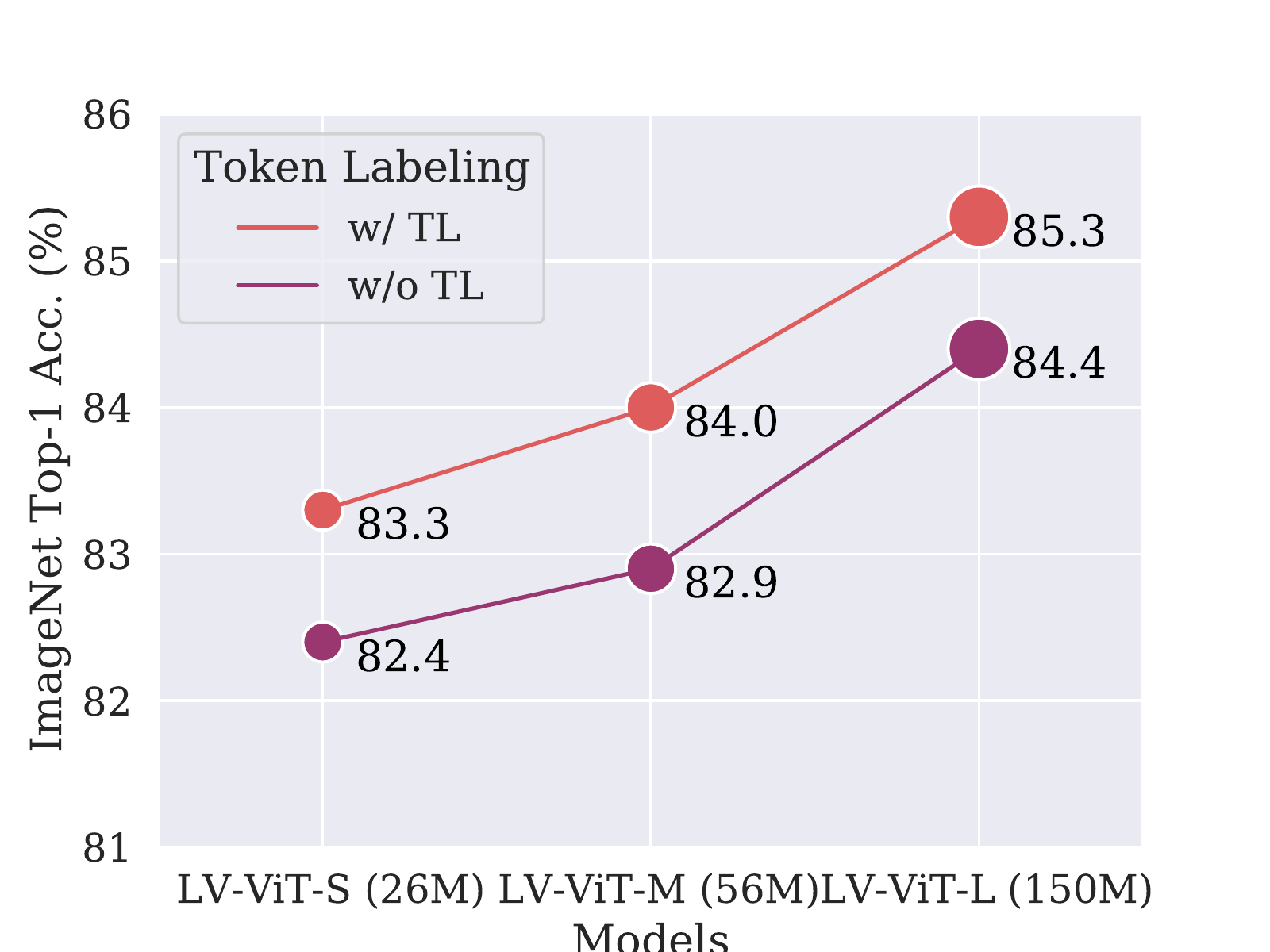}
    \end{tabular}
    \caption{Performance of the proposed token labeling objective on three different
    vision transformers: DeiT \cite{touvron2020training} (\textbf{Left}), T2T-ViT \cite{yuan2021tokens} (\textbf{Middle}), and LV-ViT (\textbf{Right}). Our method
    has a consistent improvement on all 7 different ViT models.}
    \label{fig:abl_tl}
\end{figure*}

\myPara{Robustness to Different ViT Variants} 
To further evaluate the robustness of our token labeling, we train different transformer-based networks, including DeiT~\cite{touvron2020training},
T2T-ViT~\cite{brock2021high} and our model LV-ViT, with the proposed training objective.
Results are shown in Figure~\ref{fig:abl_tl}. 
It can be found that, all the models trained with token labeling consistently outperform their vanilla counterparts, demonstrating the robustness of token labeling with respect to different variants of patch-based vision transformers.
Meanwhile, for different scales of the models, the improvement is also consistent.
Interestingly, we observe larger improvements for larger models.
These indicate that our proposed token labeling method is widely applicable to a large range of patch-based vision transformer variants.

\begin{table}[htp!]
    \centering
    \setlength\tabcolsep{1.9mm}
    \caption{Top-1 accuracy comparison with other methods on ImageNet \cite{deng2009imagenet}
    and ImageNet Real~\cite{beyer2020we}. All models are trained without external data. 
    With the same computation and parameter constraint, our model consistently outperforms
    other CNN-based and transformer-based counterparts. The results of CNNs and ViT are referenced from~\cite{touvron2021going}.}
    \label{tab:sota}
    \def \mysp {\hspace{7pt}}
    {\small 
    \begin{tabular}{@{\ }l@{\ }@{\ }lccccccc}
    \toprule
    & Network  & Params & FLOPs & Train size & Test size  &  Top-1(\%)  & Real Top-1 (\%) \\
    \toprule
    % \multicolumn{7}{c}{CNNs}\\
    % \midrule
    \multirow{6}{*}{\rotatebox{90}{CNNs}} 
    & EfficientNet-B5~\cite{tan2019efficientnet}    & \pzo30M & \dzo9.9B  & $456$ & $456$  & 83.6 & 88.3  \\
    & EfficientNet-B7~\cite{tan2019efficientnet}    & \pzo66M & \pzo37.0B & $600$ & $600$  & 84.3 & \_      \\
    & Fix-EfficientNet-B8~\cite{tan2019efficientnet, touvron2019fixing} & \pzo87M & \pzo89.5B & $672$ & $800$  & 85.7 & 90.0  \\
    % \midrule
    % & NFNet-F0~\cite{brock2021high}           & \pzo72M & \pzo12.4B & $192$ & $256$ & 83.6 & 88.1  \\
    % & NFNet-F1~\cite{brock2021high}           & 133M    & \pzo35.5B & $224$ & $320$  & 84.7 & 88.9  \\
    % & NFNet-F2~\cite{brock2021high}           & 194M    & \pzo62.6B & $256$ & $352$  & 85.1 & 88.9  \\
    & NFNet-F3~\cite{brock2021high}           & 255M    & 114.8B    & $320$ & $416$  & 85.7 & 89.4  \\
    & NFNet-F4~\cite{brock2021high}           & 316M    & 215.3B    & $384$ & $512$   & 85.9 & 89.4  \\
    & NFNet-F5~\cite{brock2021high}           & 377M    & 289.8B    & $416$ & $544$   & 86.0 & 89.2  \\
    \toprule
    % \multicolumn{7}{c}{Transformers}\\
    % \midrule
    \multirow{16}{*}{\rotatebox{90}{Transformers}} 
    & ViT-B/16~\cite{dosovitskiy2020image}           & \pzo86M & \pzo55.4B & $224$ & $384$ & 77.9 & 83.6 \\
    & ViT-L/16~\cite{dosovitskiy2020image}           & 307M    & 190.7B    & $224$ & $384$  & 76.5 & 82.2 \\
    % \midrule
    & T2T-ViT-14~\cite{yuan2021tokens}       & \pzo22M & \dzo5.2B  & $224$ & $224$        & 81.5 & \_  \\
    & T2T-ViT-14$\uparrow$384~\cite{yuan2021tokens} & \pzo22M & \pzo 17.1B  & $224$ & $384$        & 83.3 & \_ \\
    % \midrule
    & CrossViT~\cite{chen2021crossvit}           & \pzo45M & \pzo56.6B & $224$ & $480$        & 84.1 & \_ \\
    & Swin-B~\cite{liu2021swin}             & \pzo88M & \pzo47.0B   & $224$ & $384$        & 84.2 & \_ \\
    & TNT-B~\cite{han2021transformer}              & \pzo66M & \pzo14.1B & $224$ & $224$       & 82.8 & \_   \\
    % \midrule
    & DeepViT-S~\cite{zhou2021deepvit}          & \pzo27M & \dzo6.2B & $224$ & $224$       & 82.3 & \_   \\
    & DeepViT-L~\cite{zhou2021deepvit}          & \pzo55M & \pzo12.5B & $224$ & $224$       & 83.1 & \_   \\         
    % \midrule
    & DeiT-S~\cite{touvron2020training}             & \pzo22M & \dzo4.6B  & $224$ & $224$  & 79.9  & 85.7  \\
    & DeiT-B~\cite{touvron2020training}             & \pzo86M & \pzo17.5B & $224$ & $224$ &  81.8 &   86.7 \\
    & DeiT-B$\uparrow$384~\cite{touvron2020training}         & \pzo86M & \pzo55.4B & $224$  &  $384$ &  83.1 & 87.7 \\
    % \midrule
    & BoTNet-S1-128~\cite{srinivas2021bottleneck} & 79.1M & \pzo19.3B & 256 & 256 & 84.2 & - \\
    & BoTNet-S1-128$\uparrow$384~\cite{srinivas2021bottleneck} & 79.1M & \pzo45.8B & 256 & 384 & 84.7 & - \\
    & CaiT-S36$\uparrow$384~\cite{touvron2021going}  & \pzo68M & \pzo48.0B &  $224$ &  $384$ &  85.4 & 89.8 \\
    & CaiT-M36~\cite{touvron2021going} & 271M & \pzo53.7B &  $224$ &  $224$ &  85.1 & 89.3 \\
    % & CaiT-M36$\uparrow$384 & 271M & 173.3B &   $224$   & $384$ &  86.1 & 90.0 \\
    & CaiT-M36$\uparrow$448~\cite{touvron2021going} & 271M & 247.8B & $224$ & $448$  & 86.3 & 90.2\\
    \toprule
    
    % \multicolumn{7}{c}{Hybrid} \\
    % \midrule
    % \multirow{4}{*}{\rotatebox{90}{Hybrid}} 
    % & BoTNet-S1-59~\cite{srinivas2021bottleneck} & 33.5M & \dzo7.3B & 224 & 224 & 81.7 & -\\
    % & BoTNet-S1-110~\cite{srinivas2021bottleneck} & 54.7M & \pzo10.9B & 224 & 224 & 82.8 & - \\
    %  \toprule
    % \multicolumn{7}{c}{Our LV-ViT} \\
    % \midrule
    \multirow{7}{*}{\rotatebox{90}{Ours \OURS}} 
    & \OURS-S & \pzo26M & \dzo6.6B &  $224$ &  $224$  &  83.3 & 88.1 \\
    & \OURS-S$\uparrow$384 & \pzo26M & \pzo22.2B &  $224$ &  384  &  84.4 & 88.9 \\
    & \OURS-M & \pzo56M & \pzo16.0B &  $224$ &  $224$  &  84.1 & 88.4 \\
    & \OURS-M$\uparrow$384  & \pzo56M & \pzo42.2B &  $224$ &  $384$ &  85.4 & 89.5 \\
    & \OURS-L & 150M & \pzo59.0B &  $288$ &  $288$ &  85.3 & 89.3 \\
    & \OURS-L$\uparrow$448 & 150M & 157.2B &  $288$ &  $448$ &  85.9 & 89.7  \\
    & \OURS-L$\uparrow$448 & 150M & 157.2B &  $448$ &  $448$ &  86.2 & 89.9  \\
    & \OURS-L$\uparrow$512 & 151M & 214.8B &  $448$ &  $512$ &  86.4 & 90.1  \\
    \bottomrule
    \end{tabular}}
\end{table}

\subsection{Comparison to Other Methods} \label{sec:comp_others}
We compare our proposed model \nameofmethod{} with other state-of-the-art methods in Table~\ref{tab:sota}.
For small-sized models, when the test resolution is set to $224 \times 224$, 
we achieve an $83.3\%$ accuracy on ImageNet with only 26M parameters, which is $3.4\%$
higher than the strong baseline DeiT-S~\cite{touvron2020training}. 
For medium-sized models, when the test resolution is set to $384\times 384$ 
we achieve the performance of $85.4\%$, the same as CaiT-S36~\cite{touvron2021going}, but with much less computational cost and parameters. 
Note that both DeiT and CaiT use knowledge distillation to improve their models,
which introduce much more computations in training.
However, we do not require any extra computations in training and only have to compute and store the dense score maps in advance.
For large-sized models, our LV-ViT-L with a test resolution of $512\times 512$ achieves an 86.4\% top-1 accuracy, which is better than CaiT-M36~\cite{touvron2021going}  but with far fewer parameters and FLOPs.

\subsection{Semantic Segmentation on ADE20K}

It has been shown in \cite{he2019bag} that different training techniques for pretrained models have different impacts on downstream tasks with dense prediction, 
like semantic segmentation.
To demonstrate the advantage of the proposed token labeling objective on tasks with dense prediction, we apply our pretrained \nameofmethod{} with token labeling to the semantic segmentation task.

Similar to previous work \cite{liu2021swin}, we run experiments on the widely-used ADE20K \cite{zhou2019semantic} dataset. 
ADE20K contains 25K images in total, including 20K images for training, 2K images for validation and 3K images for test, and covering 150 different foreground categories.
We take both FCN \cite{long2015fully} and UperNet \cite{xiao2018unified} as our segmentation frameworks and use the mmseg toolbox to implement.
During training, following \cite{liu2021swin}, we use the AdamW optimizer with an initial learning rate of 6e-5 and a weight decay
of 0.01.
We also use a linear learning schedule with a minimum learning rate of 5e-6.
All models are trained on 8 GPUs and with a batch size of 16 (i.e., 2 images on each GPU).
The input resolution is set to $512\times512$.
In inference, a multi-scale test with interpolation rates of 
[0.75, 1.0, 1.25, 1.5, 1.75] is used.
As suggested by \cite{zhou2019semantic}, we report results in terms of both mean intersection-over-union (mIoU) and the average pixel accuracy (Pixel Acc.).

In Table~\ref{tab:seg_abl}, we test the performance of token labeling on both FCN and UperNet frameworks.
The FCN framework has a light convolutional head and can
directly reflect the performance of the pretrained models
in terms of transferable capability.
As can be seen, pretrained models with token labeling perform better than
those without token labeling.
This indicates token labeling is indeed beneficial to semantic segmentation.

We also compare our segmentation results with previous state-of-the-art 
segmentation methods in Table~\ref{tab:seg_comp}.
Without pretraining on large-scale datasets such as ImageNet-22K, our \nameofmethod{}-M with the UperNet segmentation architecture achieves an mIoU score of 50.6 with
only 77M parameters.
This result is much better than the previous CNN-based and transformer-based models.
Furthermore, using our \nameofmethod{}-L as the pretrained model yields a better
result of 51.8 in terms of mIoU.
As far as we know, this is the best result reported on ADE20K with no pretraining
on ImageNet-22K or other large-scale datasets.

\begin{table}[t]
  \centering
  \small
  \setlength\tabcolsep{1.1mm}
  \renewcommand\arraystretch{1.2}
  \caption{Transfer performance of the proposed \nameofmethod{} in semantic segmentation.
  We take two classic methods,
  FCN and UperNet, as segmentation architectures and show both single-scale (SS) and multi-scale (MS) results on the validation set.}
  \label{tab:seg_abl}
  \begin{tabular}{lcccccc} \toprule[0.5pt]
    Method & Token Labeling & Model Size & mIoU (SS) & P. Acc. (SS) & mIoU (MS) & P. Acc. (MS) \\ \midrule[0.5pt] \midrule[0.5pt]
    \nameofmethod{}-S + FCN & \xmark & 30M & 46.1 & 81.9 &  47.3 & 82.6 \\
    \nameofmethod{}-S + FCN & \cmark & 30M & 47.2 & 82.4 & 48.4 & 83.0  \\ \midrule[0.5pt]
    \nameofmethod{}-S + UperNet & \xmark & 44M & 46.5 & 82.1 &  47.6 & 82.7\\
    \nameofmethod{}-S + UperNet & \cmark & 44M & 47.9 & 82.6 & 48.6 & 83.1 \\
    \bottomrule[0.5pt]
  \end{tabular}
\end{table}

\begin{table}[h]
  \centering
  \small
  \setlength\tabcolsep{3mm}
  \renewcommand\arraystretch{1.0}
  \caption{Comparison with previous work on ADE20K validation set. 
  As far as we know, 
  our \nameofmethod{}-L + UperNet achieves the best result on ADE20K with only ImageNet-1K
  as training data in pretraining. $^\dagger$Pretrained on ImageNet-22K.}
  \label{tab:seg_comp}
  \begin{tabular}{llccccc} \toprule[0.5pt]
    & Backbone & Segmentation Architecture & Model Size & mIoU (MS) & Pixel Acc. (MS) \\ \midrule[0.5pt] \midrule[0.5pt]
    \multirow{4}{*}{\rotatebox{90}{CNNs}} 
    & ResNet-269 & PSPNet \cite{zhao2017pyramid} & - & 44.9 & 81.7 \\
    & ResNet-101 & UperNet \cite{xiao2018unified} & 86M & 44.9 & - \\
    & ResNet-101 & Strip Pooling \cite{hou2020strip} & - & 45.6 & 82.1\\
    & ResNeSt200 & DeepLabV3+ \cite{chen2018encoder} & 88M & 48.4& -  \\ 
    \midrule[0.5pt]
    \multirow{6}{*}{\rotatebox{90}{Transformers}} 
    & DeiT-S & UperNet & 52M & 44.0 & -\\
    & ViT-Large$^\dagger$ & SETR \cite{zheng2020rethinking} & 308M & 50.3 &  83.5 \\
    & Swin-T \cite{liu2021swin} & UperNet & 60M & 46.1 & - \\ 
    & Swin-S \cite{liu2021swin} & UperNet & 81M & 49.3 & - \\ 
    & Swin-B \cite{liu2021swin} & UperNet  & 121M & 49.7 & - \\
    & Swin-B$^\dagger$ \cite{liu2021swin} & UperNet  & 121M & 51.6 & - \\ \midrule[0.5pt]
    \multirow{4}{*}{\rotatebox{90}{\OURS}} 
    & \nameofmethod{}-S & FCN  & 30M & 48.4 & 83.0 \\
    & \nameofmethod{}-S & UperNet  & 44M & 48.6 & 83.1 \\
    & \nameofmethod{}-M & UperNet  & 77M & 50.6 & 83.5 \\
    & \nameofmethod{}-L & UperNet  & 209M & \highlight{51.8} & \highlight{84.1} \\
    \bottomrule[0.5pt]
  \end{tabular}
\end{table}

\section{Conclusions and Discussion} \label{sec:conclusion}

In this paper, we introduce a new token labeling method to help improve the performance  of vision transformers.
We also analyze the effectiveness and robustness of our token labeling
with respect to different annotators and different variants of patch-based vision transformers.
By applying token labeling, our proposed \nameofmethod{} achieves 84.4\% Top-1 accuracy
with only 26M parameters and 86.4\% Top-1 accuracy with 150M parameters on ImageNet-1K benchmark.

Despite the effectiveness, token labeling has a limitation of requiring a pretrained model as the machine annotator.
Fortunately, the machine annotating procedure can be done in advance to avoid introducing extra computational cost in training.
This makes our method quite different from knowledge distillation methods that rely on online teaching.
For users with limited machine resources on hand, our token labeling provides a promising training technique to improve the performance of vision transformers.

\medskip

{
\small
\bibliographystyle{plain}
\bibliography{ref}

\begin{thebibliography}{10}

\bibitem{bello2021lambdanetworks}
Irwan Bello.
\newblock Lambdanetworks: Modeling long-range interactions without attention.
\newblock {\em arXiv preprint arXiv:2102.08602}, 2021.

\bibitem{beyer2020we}
Lucas Beyer, Olivier~J H{\'e}naff, Alexander Kolesnikov, Xiaohua Zhai, and
  A{\"a}ron van~den Oord.
\newblock Are we done with imagenet?
\newblock {\em arXiv preprint arXiv:2006.07159}, 2020.

\bibitem{brock2021high}
Andrew Brock, Soham De, Samuel~L Smith, and Karen Simonyan.
\newblock High-performance large-scale image recognition without normalization.
\newblock {\em arXiv preprint arXiv:2102.06171}, 2021.

\bibitem{brown2020language}
Tom~B Brown, Benjamin Mann, Nick Ryder, Melanie Subbiah, Jared Kaplan, Prafulla
  Dhariwal, Arvind Neelakantan, Pranav Shyam, Girish Sastry, Amanda Askell,
  et~al.
\newblock Language models are few-shot learners.
\newblock {\em arXiv preprint arXiv:2005.14165}, 2020.

\bibitem{carion2020end}
Nicolas Carion, Francisco Massa, Gabriel Synnaeve, Nicolas Usunier, Alexander
  Kirillov, and Sergey Zagoruyko.
\newblock End-to-end object detection with transformers.
\newblock {\em arXiv preprint arXiv:2005.12872}, 2020.

\bibitem{chefer2020transformer}
Hila Chefer, Shir Gur, and Lior Wolf.
\newblock Transformer interpretability beyond attention visualization.
\newblock {\em arXiv preprint arXiv:2012.09838}, 2020.

\bibitem{chen2021crossvit}
Chun-Fu Chen, Quanfu Fan, and Rameswar Panda.
\newblock Crossvit: Cross-attention multi-scale vision transformer for image
  classification.
\newblock {\em arXiv preprint arXiv:2103.14899}, 2021.

\bibitem{chen2020pre}
Hanting Chen, Yunhe Wang, Tianyu Guo, Chang Xu, Yiping Deng, Zhenhua Liu, Siwei
  Ma, Chunjing Xu, Chao Xu, and Wen Gao.
\newblock Pre-trained image processing transformer.
\newblock {\em arXiv preprint arXiv:2012.00364}, 2020.

\bibitem{chen2018encoder}
Liang-Chieh Chen, Yukun Zhu, George Papandreou, Florian Schroff, and Hartwig
  Adam.
\newblock Encoder-decoder with atrous separable convolution for semantic image
  segmentation.
\newblock In {\em Proceedings of the European conference on computer vision
  (ECCV)}, pages 801--818, 2018.

\bibitem{chen2020generative}
Mark Chen, Alec Radford, Rewon Child, Jeffrey Wu, Heewoo Jun, David Luan, and
  Ilya Sutskever.
\newblock Generative pretraining from pixels.
\newblock In {\em International Conference on Machine Learning}, pages
  1691--1703. PMLR, 2020.

\bibitem{cubuk2020randaugment}
Ekin~D Cubuk, Barret Zoph, Jonathon Shlens, and Quoc~V Le.
\newblock Randaugment: Practical automated data augmentation with a reduced
  search space.
\newblock In {\em Proceedings of the IEEE/CVF Conference on Computer Vision and
  Pattern Recognition Workshops}, pages 702--703, 2020.

\bibitem{dai2020up}
Zhigang Dai, Bolun Cai, Yugeng Lin, and Junying Chen.
\newblock Up-detr: Unsupervised pre-training for object detection with
  transformers.
\newblock {\em arXiv preprint arXiv:2011.09094}, 2020.

\bibitem{d2021convit}
St{\'e}phane d'Ascoli, Hugo Touvron, Matthew Leavitt, Ari Morcos, Giulio
  Biroli, and Levent Sagun.
\newblock Convit: Improving vision transformers with soft convolutional
  inductive biases.
\newblock {\em arXiv preprint arXiv:2103.10697}, 2021.

\bibitem{deng2009imagenet}
Jia Deng, Wei Dong, Richard Socher, Li-Jia Li, Kai Li, and Li~Fei-Fei.
\newblock Imagenet: A large-scale hierarchical image database.
\newblock In {\em 2009 IEEE conference on computer vision and pattern
  recognition}, pages 248--255. Ieee, 2009.

\bibitem{devlin2018bert}
Jacob Devlin, Ming-Wei Chang, Kenton Lee, and Kristina Toutanova.
\newblock Bert: Pre-training of deep bidirectional transformers for language
  understanding.
\newblock {\em arXiv preprint arXiv:1810.04805}, 2018.

\bibitem{dosovitskiy2020image}
Alexey Dosovitskiy, Lucas Beyer, Alexander Kolesnikov, Dirk Weissenborn,
  Xiaohua Zhai, Thomas Unterthiner, Mostafa Dehghani, Matthias Minderer, Georg
  Heigold, Sylvain Gelly, et~al.
\newblock An image is worth 16x16 words: Transformers for image recognition at
  scale.
\newblock {\em arXiv preprint arXiv:2010.11929}, 2020.

\bibitem{han2021transformer}
Kai Han, An~Xiao, Enhua Wu, Jianyuan Guo, Chunjing Xu, and Yunhe Wang.
\newblock Transformer in transformer.
\newblock {\em arXiv preprint arXiv:2103.00112}, 2021.

\bibitem{he2017mask}
Kaiming He, Georgia Gkioxari, Piotr Doll{\'a}r, and Ross Girshick.
\newblock Mask r-cnn.
\newblock In {\em Proceedings of the IEEE international conference on computer
  vision}, pages 2961--2969, 2017.

\bibitem{he2016deep}
Kaiming He, Xiangyu Zhang, Shaoqing Ren, and Jian Sun.
\newblock Deep residual learning for image recognition.
\newblock In {\em Proceedings of the IEEE conference on computer vision and
  pattern recognition}, pages 770--778, 2016.

\bibitem{he2019bag}
Tong He, Zhi Zhang, Hang Zhang, Zhongyue Zhang, Junyuan Xie, and Mu~Li.
\newblock Bag of tricks for image classification with convolutional neural
  networks.
\newblock In {\em Proceedings of the IEEE/CVF Conference on Computer Vision and
  Pattern Recognition}, pages 558--567, 2019.

\bibitem{heo2021rethinking}
Byeongho Heo, Sangdoo Yun, Dongyoon Han, Sanghyuk Chun, Junsuk Choe, and
  Seong~Joon Oh.
\newblock Rethinking spatial dimensions of vision transformers.
\newblock {\em arXiv preprint arXiv:2103.16302}, 2021.

\bibitem{hinton2015distilling}
Geoffrey Hinton, Oriol Vinyals, and Jeff Dean.
\newblock Distilling the knowledge in a neural network.
\newblock {\em arXiv preprint arXiv:1503.02531}, 2015.

\bibitem{hou2020strip}
Qibin Hou, Li~Zhang, Ming-Ming Cheng, and Jiashi Feng.
\newblock Strip pooling: Rethinking spatial pooling for scene parsing.
\newblock In {\em Proceedings of the IEEE/CVF Conference on Computer Vision and
  Pattern Recognition}, pages 4003--4012, 2020.

\bibitem{huang2016deep}
Gao Huang, Yu~Sun, Zhuang Liu, Daniel Sedra, and Kilian~Q Weinberger.
\newblock Deep networks with stochastic depth.
\newblock In {\em European conference on computer vision}, pages 646--661.
  Springer, 2016.

\bibitem{liu2019roberta}
Yinhan Liu, Myle Ott, Naman Goyal, Jingfei Du, Mandar Joshi, Danqi Chen, Omer
  Levy, Mike Lewis, Luke Zettlemoyer, and Veselin Stoyanov.
\newblock Roberta: A robustly optimized bert pretraining approach.
\newblock {\em arXiv preprint arXiv:1907.11692}, 2019.

\bibitem{liu2021swin}
Ze~Liu, Yutong Lin, Yue Cao, Han Hu, Yixuan Wei, Zheng Zhang, Stephen Lin, and
  Baining Guo.
\newblock Swin transformer: Hierarchical vision transformer using shifted
  windows.
\newblock {\em arXiv preprint arXiv:2103.14030}, 2021.

\bibitem{long2015fully}
Jonathan Long, Evan Shelhamer, and Trevor Darrell.
\newblock Fully convolutional networks for semantic segmentation.
\newblock In {\em Proceedings of the IEEE conference on computer vision and
  pattern recognition}, pages 3431--3440, 2015.

\bibitem{loshchilov2017decoupled}
Ilya Loshchilov and Frank Hutter.
\newblock Decoupled weight decay regularization.
\newblock {\em arXiv preprint arXiv:1711.05101}, 2017.

\bibitem{parmar2018image}
Niki Parmar, Ashish Vaswani, Jakob Uszkoreit, {\L}ukasz Kaiser, Noam Shazeer,
  Alexander Ku, and Dustin Tran.
\newblock Image transformer.
\newblock {\em arXiv preprint arXiv:1802.05751}, 2018.

\bibitem{paszke2019pytorch}
Adam Paszke, Sam Gross, Francisco Massa, Adam Lerer, James Bradbury, Gregory
  Chanan, Trevor Killeen, Zeming Lin, Natalia Gimelshein, Luca Antiga, et~al.
\newblock Pytorch: An imperative style, high-performance deep learning library.
\newblock In {\em Advances in neural information processing systems}, pages
  8026--8037, 2019.

\bibitem{srinivas2021bottleneck}
Aravind Srinivas, Tsung-Yi Lin, Niki Parmar, Jonathon Shlens, Pieter Abbeel,
  and Ashish Vaswani.
\newblock Bottleneck transformers for visual recognition.
\newblock {\em arXiv preprint arXiv:2101.11605}, 2021.

\bibitem{srivastava2014dropout}
Nitish Srivastava, Geoffrey Hinton, Alex Krizhevsky, Ilya Sutskever, and Ruslan
  Salakhutdinov.
\newblock Dropout: a simple way to prevent neural networks from overfitting.
\newblock {\em The journal of machine learning research}, 15(1):1929--1958,
  2014.

\bibitem{sun2020rethinking}
Zhiqing Sun, Shengcao Cao, Yiming Yang, and Kris Kitani.
\newblock Rethinking transformer-based set prediction for object detection.
\newblock {\em arXiv preprint arXiv:2011.10881}, 2020.

\bibitem{tan2019efficientnet}
Mingxing Tan and Quoc~V Le.
\newblock Efficientnet: Rethinking model scaling for convolutional neural
  networks.
\newblock {\em arXiv preprint arXiv:1905.11946}, 2019.

\bibitem{tolstikhin2021mlp}
Ilya Tolstikhin, Neil Houlsby, Alexander Kolesnikov, Lucas Beyer, Xiaohua Zhai,
  Thomas Unterthiner, Jessica Yung, Daniel Keysers, Jakob Uszkoreit, Mario
  Lucic, et~al.
\newblock Mlp-mixer: An all-mlp architecture for vision.
\newblock {\em arXiv preprint arXiv:2105.01601}, 2021.

\bibitem{touvron2020training}
Hugo Touvron, Matthieu Cord, Matthijs Douze, Francisco Massa, Alexandre
  Sablayrolles, and Herv{\'e} J{\'e}gou.
\newblock Training data-efficient image transformers \& distillation through
  attention.
\newblock {\em arXiv preprint arXiv:2012.12877}, 2020.

\bibitem{touvron2021going}
Hugo Touvron, Matthieu Cord, Alexandre Sablayrolles, Gabriel Synnaeve, and
  Herv{\'e} J{\'e}gou.
\newblock Going deeper with image transformers.
\newblock {\em arXiv preprint arXiv:2103.17239}, 2021.

\bibitem{touvron2019fixing}
Hugo Touvron, Andrea Vedaldi, Matthijs Douze, and Herv{\'e} J{\'e}gou.
\newblock Fixing the train-test resolution discrepancy.
\newblock {\em arXiv preprint arXiv:1906.06423}, 2019.

\bibitem{vaswani2017attention}
Ashish Vaswani, Noam Shazeer, Niki Parmar, Jakob Uszkoreit, Llion Jones,
  Aidan~N Gomez, {\L}ukasz Kaiser, and Illia Polosukhin.
\newblock Attention is all you need.
\newblock {\em Advances in neural information processing systems},
  30:5998--6008, 2017.

\bibitem{wang2021pyramid}
Wenhai Wang, Enze Xie, Xiang Li, Deng-Ping Fan, Kaitao Song, Ding Liang, Tong
  Lu, Ping Luo, and Ling Shao.
\newblock Pyramid vision transformer: A versatile backbone for dense prediction
  without convolutions.
\newblock {\em arXiv preprint arXiv:2102.12122}, 2021.

\bibitem{wang2020end}
Yuqing Wang, Zhaoliang Xu, Xinlong Wang, Chunhua Shen, Baoshan Cheng, Hao Shen,
  and Huaxia Xia.
\newblock End-to-end video instance segmentation with transformers.
\newblock {\em arXiv preprint arXiv:2011.14503}, 2020.

\bibitem{rw2019timm}
Ross Wightman.
\newblock Pytorch image models.
\newblock \url{https://github.com/rwightman/pytorch-image-models}, 2019.

\bibitem{wu2021cvt}
Haiping Wu, Bin Xiao, Noel Codella, Mengchen Liu, Xiyang Dai, Lu~Yuan, and Lei
  Zhang.
\newblock Cvt: Introducing convolutions to vision transformers.
\newblock {\em arXiv preprint arXiv:2103.15808}, 2021.

\bibitem{xiao2018unified}
Tete Xiao, Yingcheng Liu, Bolei Zhou, Yuning Jiang, and Jian Sun.
\newblock Unified perceptual parsing for scene understanding.
\newblock In {\em Proceedings of the European Conference on Computer Vision
  (ECCV)}, pages 418--434, 2018.

\bibitem{yang2020learning}
Fuzhi Yang, Huan Yang, Jianlong Fu, Hongtao Lu, and Baining Guo.
\newblock Learning texture transformer network for image super-resolution.
\newblock In {\em Proceedings of the IEEE/CVF Conference on Computer Vision and
  Pattern Recognition}, pages 5791--5800, 2020.

\bibitem{yuan2021tokens}
Li~Yuan, Yunpeng Chen, Tao Wang, Weihao Yu, Yujun Shi, Francis~EH Tay, Jiashi
  Feng, and Shuicheng Yan.
\newblock Tokens-to-token vit: Training vision transformers from scratch on
  imagenet.
\newblock {\em arXiv preprint arXiv:2101.11986}, 2021.

\bibitem{yuan2020revisiting}
Li~Yuan, Francis~EH Tay, Guilin Li, Tao Wang, and Jiashi Feng.
\newblock Revisiting knowledge distillation via label smoothing regularization.
\newblock In {\em Proceedings of the IEEE/CVF Conference on Computer Vision and
  Pattern Recognition}, pages 3903--3911, 2020.

\bibitem{yun2019cutmix}
Sangdoo Yun, Dongyoon Han, Seong~Joon Oh, Sanghyuk Chun, Junsuk Choe, and
  Youngjoon Yoo.
\newblock Cutmix: Regularization strategy to train strong classifiers with
  localizable features.
\newblock In {\em Proceedings of the IEEE/CVF International Conference on
  Computer Vision}, pages 6023--6032, 2019.

\bibitem{yun2021relabel}
Sangdoo Yun, Seong~Joon Oh, Byeongho Heo, Dongyoon Han, Junsuk Choe, and
  Sanghyuk Chun.
\newblock Re-labeling imagenet: from single to multi-labels, from global to
  localized labels.
\newblock {\em arXiv preprint arXiv:2101.05022}, 2021.

\bibitem{zeng2020learning}
Yanhong Zeng, Jianlong Fu, and Hongyang Chao.
\newblock Learning joint spatial-temporal transformations for video inpainting.
\newblock In {\em European Conference on Computer Vision}, pages 528--543.
  Springer, 2020.

\bibitem{zhang2020resnest}
Hang Zhang, Chongruo Wu, Zhongyue Zhang, Yi~Zhu, Zhi Zhang, Haibin Lin, Yue
  Sun, Tong He, Jonas Muller, R.~Manmatha, Mu~Li, and Alexander Smola.
\newblock Resnest: Split-attention networks.
\newblock {\em arXiv preprint arXiv:2004.08955}, 2020.

\bibitem{zhang2017mixup}
Hongyi Zhang, Moustapha Cisse, Yann~N Dauphin, and David Lopez-Paz.
\newblock mixup: Beyond empirical risk minimization.
\newblock {\em arXiv preprint arXiv:1710.09412}, 2017.

\bibitem{zhao2020point}
Hengshuang Zhao, Li~Jiang, Jiaya Jia, Philip Torr, and Vladlen Koltun.
\newblock Point transformer.
\newblock {\em arXiv preprint arXiv:2012.09164}, 2020.

\bibitem{zhao2017pyramid}
Hengshuang Zhao, Jianping Shi, Xiaojuan Qi, Xiaogang Wang, and Jiaya Jia.
\newblock Pyramid scene parsing network.
\newblock In {\em Proceedings of the IEEE conference on computer vision and
  pattern recognition}, pages 2881--2890, 2017.

\bibitem{zheng2020end}
Minghang Zheng, Peng Gao, Xiaogang Wang, Hongsheng Li, and Hao Dong.
\newblock End-to-end object detection with adaptive clustering transformer.
\newblock {\em arXiv preprint arXiv:2011.09315}, 2020.

\bibitem{zheng2020rethinking}
Sixiao Zheng, Jiachen Lu, Hengshuang Zhao, Xiatian Zhu, Zekun Luo, Yabiao Wang,
  Yanwei Fu, Jianfeng Feng, Tao Xiang, Philip~HS Torr, et~al.
\newblock Rethinking semantic segmentation from a sequence-to-sequence
  perspective with transformers.
\newblock {\em arXiv preprint arXiv:2012.15840}, 2020.

\bibitem{zhong2020random}
Zhun Zhong, Liang Zheng, Guoliang Kang, Shaozi Li, and Yi~Yang.
\newblock Random erasing data augmentation.
\newblock In {\em Proceedings of the AAAI Conference on Artificial
  Intelligence}, volume~34, pages 13001--13008, 2020.

\bibitem{zhou2019semantic}
Bolei Zhou, Hang Zhao, Xavier Puig, Tete Xiao, Sanja Fidler, Adela Barriuso,
  and Antonio Torralba.
\newblock Semantic understanding of scenes through the ade20k dataset.
\newblock {\em International Journal of Computer Vision}, 127(3):302--321,
  2019.

\bibitem{zhou2021deepvit}
Daquan Zhou, Bingyi Kang, Xiaojie Jin, Linjie Yang, Xiaochen Lian, Qibin Hou,
  and Jiashi Feng.
\newblock Deepvit: Towards deeper vision transformer.
\newblock {\em arXiv preprint arXiv:2103.11886}, 2021.

\bibitem{zhou2018end}
Luowei Zhou, Yingbo Zhou, Jason~J Corso, Richard Socher, and Caiming Xiong.
\newblock End-to-end dense video captioning with masked transformer.
\newblock In {\em Proceedings of the IEEE Conference on Computer Vision and
  Pattern Recognition}, pages 8739--8748, 2018.

\bibitem{zhu2020deformable}
Xizhou Zhu, Weijie Su, Lewei Lu, Bin Li, Xiaogang Wang, and Jifeng Dai.
\newblock Deformable detr: Deformable transformers for end-to-end object
  detection.
\newblock {\em arXiv preprint arXiv:2010.04159}, 2020.

\end{thebibliography}
}
\newpage
\appendix

\section{More Experiment Details}
We show the default hyper-parameters for our ImageNet classification experiments in Table~\ref{tab:hyper_p}. In addition, for fine-tuning on larger image resolution, we set batch size to 512, learning rate to 5e-6, weight decay to 1e-8 and fine-tune 30 epochs. Other hyper-parameters are set the same as default. During training, a machine node with 8 NVIDIA V100 GPUs (32G memory) is required. When fine-tuning our large model with image resolution of $448\times 448$, we need 4 machine nodes with the same GPU settings as above.

\begin{table}[h]
  \centering
  \small
  \setlength\tabcolsep{1.2mm}
  \renewcommand\arraystretch{1}
  \caption{Default hyper-parameters for our experiments. Note that we do not use the MixUp augmentation method
  when ReLabel or token labeling is used.}
  \label{tab:hyper_p}
  \begin{tabular}{lccccc} \toprule[0.5pt]
    Supervision & Standard & ReLabel & Token labeling \\ \midrule[0.5pt] \midrule[0.5pt]
    Epoch & 300 & 300 & 300\\ \midrule[0.5pt]
    Batch size & 1024 & 1024 & 1024\\
    LR & 1e-3 $\cdot \frac{\text{batch\_size}}{\text{1024}}$ &  1e-3$\cdot \frac{\text{batch\_size}}{\text{1024}}$  &  1e-3$\cdot \frac{\text{batch\_size}}{\text{640}}$ \\
    LR decay & cosine & cosine & cosine\\
    Weight decay & 0.05& 0.05& 0.05\\
    Warmup epochs& 5 & 5 & 5 \\\midrule[0.5pt]
    Dropout & 0 & 0& 0\\
    Stoch. Depth & 0.1 & 0.1 & 0.1 \\
    MixUp alpha & 0.8 & - & -\\
    Erasing prob. & 0.25 & 0.25 &0.25\\
    RandAug & 9/0.5 & 9/0.5 & 9/0.5\\
    \bottomrule[0.5pt]
  \end{tabular}
\end{table}

\section{More Experiments}

\subsection{Training Technique Analysis} \label{sec:analysis}

We present a summary of our modification and proposed token labeling method to improve vision transformer models in Table~\ref{tab:tricks}.
We take the DeiT-Small \cite{touvron2020training} model as our baseline and show the performance increment as more training techniqeus are added.
In this subsection, we will ablate the proposed modifications and evaluate the effectiveness of them. 
\begin{table}[h]
  \centering
  \small
  \setlength\tabcolsep{1mm}
  \renewcommand\arraystretch{1}
  \caption{Ablation path from the DeiT-Small \cite{touvron2020training} baseline to our \nameofmethod{}-S.
  All experiments expect for larger input resolution can be finished within 3 days using a single server node
  with 8 V100 GPUs. Clearly, with only 26M learnable parameters, the performance can be boosted from 79.9
  to 84.4 (\highlight{+4.5}) using the proposed Token Labeling and other proposed training techniques.}
  \label{tab:tricks}
  \begin{tabular}{lccccc} \toprule[0.5pt]
    Training techniques & \#Param. & Top-1 Acc. (\%) \\ \midrule[0.5pt] \midrule[0.5pt]
    Baseline (DeiT-Small \cite{touvron2020training}) &22M &  79.9\\
    + More transformers ($12 \rightarrow 16$) & 28M & 81.2 (\highlight{+1.2}) \\
    + Less MLP expansion ratio ($4 \rightarrow 3$) & 25M & 81.1 (\highlight{+1.1})\\
    + More convs for patch embedding &26M& 82.2  (\highlight{+2.3})\\
    + Enhanced residual connection &26M& 82.4
    (\highlight{+2.5}) \\
    + Token labeling with MixToken &26M & 83.3
    (\highlight{+3.4})\\
    + Input resolution ($224 \rightarrow 384$) & 26M & 84.4 (\highlight{+4.5})\\
    \bottomrule[0.5pt]
  \end{tabular}
\end{table}

\myPara{Explicit inductive bias for patch embedding}
Ablation analysis of patch embedding is presented in Table~\ref{tab:abl_conv}. 
The baseline is set to the same as the setting as presented in the third row of Table~\ref{tab:tricks}.
Clearly, by adding more convolutional layers and narrow the kernel size in the patch embedding,
we can see a consistent increase in the performance comparing to the original single-layer patch embedding.
However, when further increasing the number of convolutional layer in patch embedding to 6, 
we do not observe any performance gain.
This indicates that using 4-layer convolutions in patch embedding is enough.
Meanwhile, if we use a larger stride to reduce the size of the feature map,
we can largely reduce the computation cost, but the performance also drops. 
Thus, we only apply a convolution of stride 2 and kernel size 7 at the beginning of 
the patch embedding module, followed by two convolutional layers with stride 1 and kernel size 3.
The feature map is finally tokenized to a sequence of tokens using a convolutional layer of stride 8 
and kernel size 8 (see the fifth line in Table~\ref{tab:abl_conv}).

\begin{table}[h]
  \centering
  \small
  \setlength\tabcolsep{1.1mm}
  \renewcommand\arraystretch{1.1}
  \caption{Ablation on patch embedding. Baseline is set as 16 layer ViT with embedding size 384 and MLP expansion ratio of 3.
  All convolutional layers except the last block have 64 filters. \#Convs indicatie the total number of convolutions for patch embedding, while the kernel size and stride correspond to each layer are shown as a list in the table.}
  \label{tab:abl_conv}
  \begin{tabular}{cccccc} \toprule[0.5pt]
    \#Convs & Kerenl size & Stride & Params & Top-1 Acc. (\%) \\ \midrule[0.5pt] \midrule[0.5pt]
    1 & [16] & [16] &25M &  81.1\\
    2 & [7,8] & [2,8] &25M &  81.4\\ 
    3 & [7,3,8] & [2,2,4] &25M &  81.4\\
    3 & [7,3,8] & [2,1,8] &26M & 81.9\\
    4 & [7,3,3,8] & [2,1,1,8] &26M & \highlight{82.2}\\
    6 & [7,3,3,3,3,8] & [2,1,1,1,1,8] &26M & \highlight{82.2}\\

    \bottomrule[0.5pt]
  \end{tabular}
\end{table}

\myPara{Enhanced residual connection}
We found that introducing a residual scaling factor can also bring benefit 
as shown in Table~\ref{tab:abl_res}.
We found that using smaller scaling factor can lead to better performance and faster convergence.
Part of the reason is that more information can be preserved in the main branch, leading to less information loss and better performance.

\begin{table}[h]
  \centering
  \small
  \setlength\tabcolsep{2.6mm}
  \renewcommand\arraystretch{1}
  \caption{Ablation on enhancing residual connection by applying a scaling factor. 
  Baseline is a 16-layer vision transformer with 4-layer convolutional patch embedding. 
  Here, function $F$ represents either self-attention (SA) or feed forward (FF).}
  
  \label{tab:abl_res}
  \begin{tabular}{lccccc} \toprule[0.5pt]
    &Forward Function & \#Parameters & Top-1 Acc. (\%) \\ \midrule[0.5pt] \midrule[0.5pt]
      &  $X \longleftarrow X +F(X)$ &26M &  82.2\\
     & $X \longleftarrow X +F(X)/2$ &26M &  \highlight{\textbf{82.4}}\\ 
     & $X \longleftarrow X +F(X)/3$ &26M &  \highlight{\textbf{82.4}}\\

    \bottomrule[0.5pt]
  \end{tabular}
\end{table}

\myPara{Larger input resolution}
To adapt our model to larger input image, we interpolate the positional encoding and fine-tune the model on larger image resolution for a few epochs. Token labeling objective as well as MixToken are also used during fine-tuning. As can be seen from Table~\ref{tab:tricks}, fine-tuning on larger input resolution of $384\times 384$ can improve the performance by $1.1\%$ for our LV-ViT-S model.

\subsection{Beyond Vision Transformers: Performance on MLP-Based and CNN-Based Models}

We further explore the performance of token labeling on other CNN-based and MLP-based models. Results are shown in Table~\ref{tab:abl_more}. Besides our re-implementation with more data augmentation and regularization technique, we also list the results from the original papers. It shows that for both MLP-based and CNN-based models, token labeling objective can still improve the performance over the strong baselines by providing the location-specific dense supervision. 

\begin{table}[H]
 \centering
 \small
 \setlength\tabcolsep{1.4mm}
 \renewcommand\arraystretch{1.1}
 \caption{Performance of the proposed token labeling objective on representative CNN-based (ResNeSt) and MLP-based (Mixer-MLP) models. Our method has a consistent improvement on all different models. Here $^\dag$ indicates results reported in original papers.}
  
 \label{tab:abl_more}
  \begin{tabular}{c|ccc|ccc|ccc|ccccccccc} \toprule[0.5pt]
    \multicolumn{1}{c}{Model}& \multicolumn{3}{c}{Mixer-S/16~\cite{tolstikhin2021mlp}} & \multicolumn{3}{c}{Mixer-B/16~\cite{tolstikhin2021mlp}} & \multicolumn{3}{c}{Mixer-L/16~\cite{tolstikhin2021mlp}} & \multicolumn{3}{c}{ResNeSt-50~\cite{zhang2020resnest}}  \\ \midrule[0.5pt] \midrule[0.5pt]
    Token Labeling & \xmark & \xmark & \checkmark & \xmark& \xmark & \checkmark& \xmark & \xmark & \checkmark & \xmark& \xmark & \checkmark \\
    Parameters & 18M& 18M & 18M & 59M&  59M & 59M & 207M& 207M & 207M & 27M& 27M & 27M \\
    Top-1 Acc. (\%) & 73.8$^\dag$& 75.6 & \textbf{76.1} & 76.4$^\dag$& 78.3 & \textbf{79.5}  & 71.6$^\dag$& 77.7 & \textbf{80.1} & 81.1$^\dag$& 80.9 & \textbf{81.5} \\

    \bottomrule[0.5pt]
  \end{tabular}
\end{table}

\subsection{Comparison with CaiT}

CaiT~\cite{touvron2021going} is currently the best transformer-based model.
We list the comparison of training hyper-parameters and model configuration with CaiT in Table~\ref{tab:comp_cait}. It can be seen that using less training techniques, computations, and smaller model size, our LV-ViT achieves identical result to the state-of-the-art CaiT model.

\begin{table}[H]
  \centering
  \small
  \setlength\tabcolsep{2.6mm}
  \renewcommand\arraystretch{1}
  \caption{Comparison with CaiT \cite{touvron2021going}. Our model exploits less training techniques,
  model size, and computations but achieve identical result to CaiT.}
  \label{tab:comp_cait}
  \begin{tabular}{lcc} \toprule[0.5pt]
    Settings & LV-ViT (Ours) & CaiT~\cite{touvron2021going} \\ \midrule[0.5pt] 
    Transformer Blocks    & 20   & 36 \\
    \#Head in Self-attention & 8 & 12 \\
    MLP Expansion Ratio & 3 & 4 \\
    Embedding Dimension & 512 & 384 \\
    Stochastic Depth \cite{huang2016deep} & 0.2 (Linear) & 0.2 (Fixed) \\
    Rand Augmentation \cite{cubuk2020randaugment} & \cmarkgreen & \cmarkgreen \\
    CutMix Augmentation \cite{yun2019cutmix} &  & \cmarkgreen \\
    MixUp Augmentation \cite{zhang2017mixup} & & \cmarkgreen \\
    LayerScaling \cite{touvron2021going} & & \cmarkgreen \\
    Class Attention \cite{touvron2021going} & & \cmarkgreen \\
    Knowledge Distillation & & \cmarkgreen \\
    Enhanced Residuals (Ours) & \cmarkgreen & \\
    MixToken (Ours) & \cmarkgreen & \\
    Token Labeling (Ours) & \cmarkgreen & \\ \midrule[0.5pt]
    Test Resolution & $384 \times 384$ & $384 \times 384$ \\
    Model Size & 56M & 69M \\
    Computations & 42B & 48B \\
    Training Epoch & 300 & 400 \\ \midrule[0.5pt]
    ImageNet Top-1 Acc. & 85.4 & 85.4 \\
    \bottomrule[0.5pt]
  \end{tabular}
\end{table}

\section{Visualization} \label{sec:visualize}

\begin{figure}[H]
    \centering
    \small
    \begin{overpic}[width=0.8\linewidth]{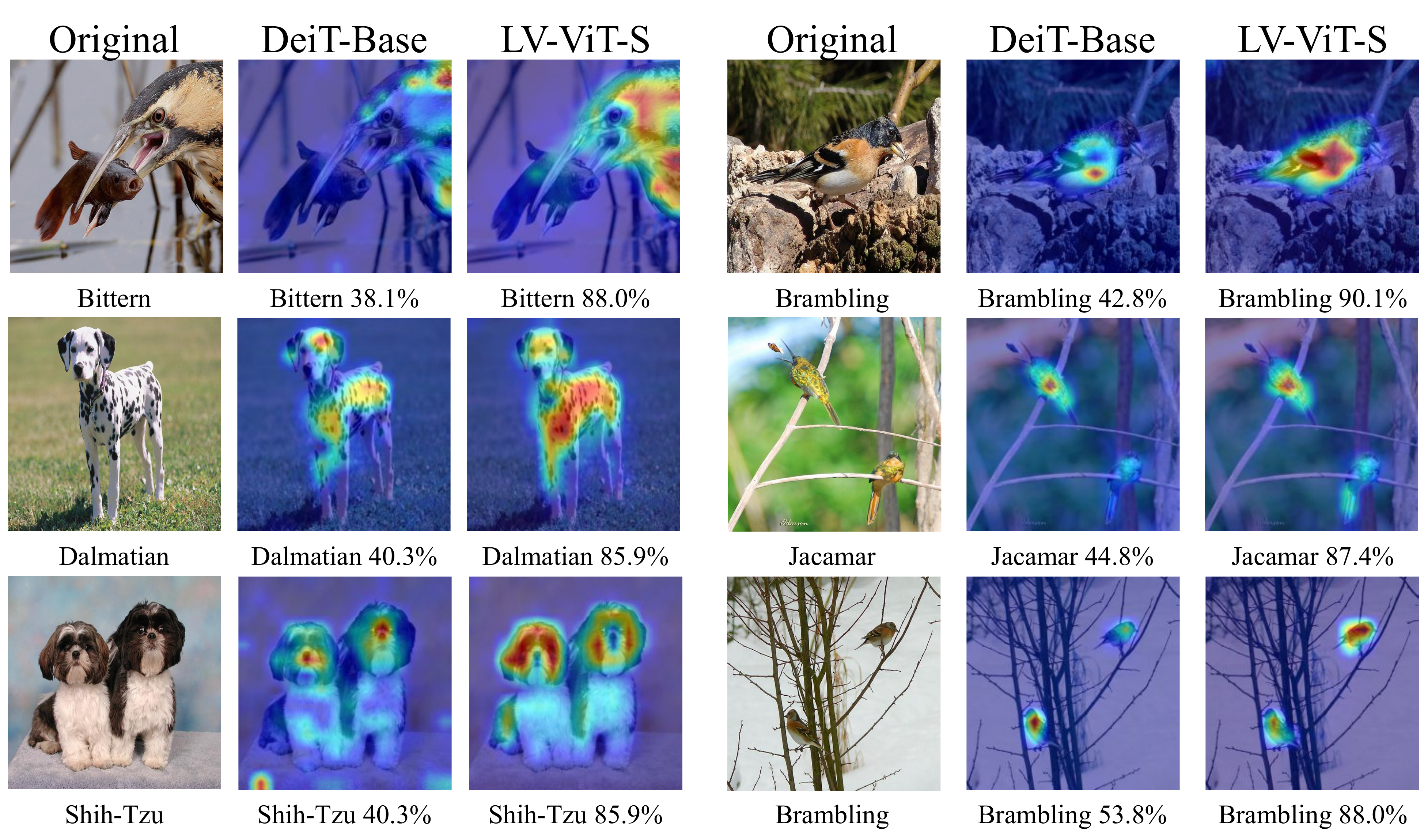}
    \end{overpic}
    \caption{Visual comparisons between DeiT-base and LV-ViT-S.}
    \label{fig:vis1}
    \vspace{-20pt}
\end{figure}

\begin{figure}[H]
    \centering
    \small
    \begin{overpic}[width=0.7\linewidth]{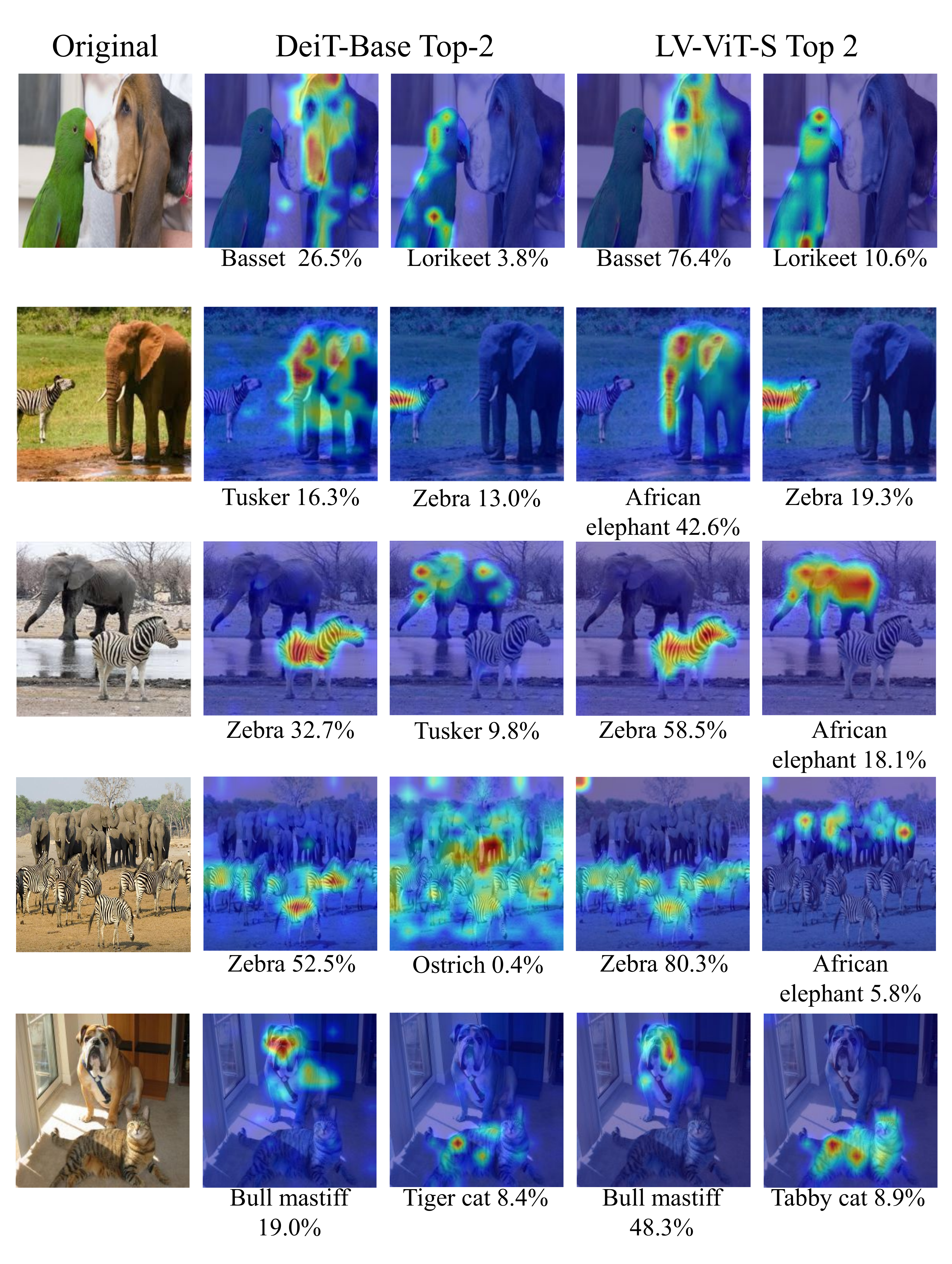}
    \end{overpic}
    \caption{Visual comparisons between DeiT-base and LV-ViT-S for the top-2 predicted classes.}
    \label{fig:vis2}
\end{figure}

We apply the method proposed in~\cite{chefer2020transformer} to visualize both DeiT-base and our LV-ViT-S.
Results are shown in Figure~\ref{fig:vis1} and Figure~\ref{fig:vis2}.
In Figure~\ref{fig:vis1}, we can observe that our LV-ViT-S model performs better in locating the target objects
and hence yields better classification performance with high confidence.  
In Figure~\ref{fig:vis2}, we visualize the top-2 classes predicted by the two models. 
Noted that we follow~\cite{chefer2020transformer} to select images with at least 2 classes existing.
It can be seen that our LV-ViT-S trained with token labeling can accurately locate both classes while the DeiT-base sometimes fails in locating the entire target object for a certain class.
This demonstrates that our token labeling objective does help in improving models' visual grounding capability
because of the location-specific token-level information.

\end{document}